\documentclass[letterpaper]{article} 
\usepackage[preprint]{aaai2027}  
\usepackage[hyphens]{url}  
\usepackage{graphicx} 
\usepackage{natbib}  
\usepackage{caption} 
\usepackage{hyperref}
\usepackage{algorithm}
\usepackage{algorithmic}
\usepackage{amsfonts}
\usepackage{amsmath}
\usepackage{amssymb}
\usepackage[table]{xcolor}
\usepackage{multirow}
\usepackage{makecell} 

\definecolor{rowOdd}{gray}{0.90}   
\definecolor{rowEven}{gray}{0.98}  
\definecolor{highlight}{gray}{0.85} 

\DeclareMathOperator{\tr}{tr}
\DeclareMathOperator{\nll}{nll}
\DeclareMathOperator{\SG}{\texttt{\textbf{SG}}}

\usepackage{newfloat}
\usepackage{listings}
\DeclareCaptionStyle{ruled}{labelfont=normalfont,labelsep=colon,strut=off} 
\floatstyle{ruled}
\newfloat{listing}{tb}{lst}{}
\floatname{listing}{Listing}

\usepackage{booktabs}

\title{NAE: Normalizing AutoEncoder}
\author {
    Muhammad Abdur Rafae\corresponding,
    Niels Landwehr
}
\affiliations {
    University of Hildeshiem \\
    rafaem/landwehr@uni-hildesheim.de
}

\begin{document}
\maketitle

\begin{abstract}
	We consider the setting of Normalizing flows with approximate inverses—an established paradigm spanning both full-dimensional ($d=D$) and bottleneck ($d<D$) settings—and group these models under the term \textit{flow autoencoders}. We present a theoretical investigation into their training dynamics and prove that the proposed loss used by existing approaches is suboptimal; specifically, both encoder and decoder surrogates must be optimized in alignment with reconstruction loss. Guided by these insights, we propose Normalizing Autoencoder (NAE), which employs a novel conditional loss that aligns the surrogate loss gradient with that of reconstruction loss, directly improving upon the current standard. Extensive experiments across molecule generation, tabular data, and image benchmarks demonstrate that NAE achieves state-of-the-art performance. Our work highlights the importance of loss alignment in flow autoencoders and establishes NAE as a powerful generative framework. 
\end{abstract}

\section{Introduction}

Normalizing flows are a powerful class of generative models that
combine tractable likelihood evaluation with flexible density
estimation by learning an invertible transformation between a simple
base distribution and a complex data distribution~\cite{walton2023}.
However, exact invertibility typically requires specialized
architectures, such as coupling layers, autoregressive
transformations, or continuous-time dynamics. These restrictions can
limit architectural flexibility or introduce costly numerical
integration during training and sampling.

Recent work relaxes these restrictions by parameterizing the forward
and inverse mappings as separate encoder and decoder
networks~\cite{sorrenson2024,draxler2024,sorrenson2024b}. The resulting
models combine the structure of autoencoders with the
likelihood-based training of normalizing flows; we refer to this
emerging model class as \textbf{flow autoencoders}. This paradigm
encompasses both full-dimensional ($d=D$) and bottleneck ($d<D$)
settings, with both settings relying on surrogate-gradient formulations for efficient likelihood optimization.

Despite their promising empirical performance, the training dynamics
of flow autoencoders remain underexplored. Prior work introduced
two surrogate formulations for estimating log-determinant gradients:
an encoder surrogate (which updates the encoder) and a decoder surrogate (which updates the decoder). Empirically, training with the encoder
surrogate is stable, whereas the decoder surrogate often leads to
instability. This raises two central questions: \emph{Why do the
	surrogates exhibit different training behavior, and can we effectively leverage both surrogates?}

In this work, we address these questions by introducing the \textbf{Normalizing Autoencoder (NAE)}, a flexible generative framework built on the flow autoencoder paradigm. Our contributions are threefold:
\begin{itemize}
	\item We analyze the training dynamics of flow autoencoders and demonstrate that both surrogates are essential for effective learning.
	\item We propose a conditional surrogate loss that dynamically selects the surrogate whose gradient is aligned with the gradient of the reconstruction loss.
	\item We evaluate NAE across graph-structured, tabular, and vision benchmarks, demonstrating state-of-the-art performance across three data modalities. 
\end{itemize}

\section{Related Works}

\paragraph{Normalizing Flows.}
Traditional normalizing flows rely on invertible transformations with tractable Jacobian determinants. Early methods used coupling and autoregressive layers with triangular Jacobians, such as NICE~\cite{dinh2015}, RealNVP~\cite{dinh2017}, MAF~\cite{papamakarios2017a}, and Glow~\cite{kingma2018}. Subsequent advances introduced spline-based coupling functions~\cite{durkan2019, hong2023}, invertible residual networks with Lipschitz constraints~\cite{behrmann2019, chen2020}, and continuous normalizing flows via neural ODEs~\cite{grathwohl2018}. While expressive, these approaches impose rigid architectural constraints or require costly numerical integration during sampling, limiting their scalability to high-dimensional problems.

\paragraph{Injective Flows.}
A limitation of standard normalizing flows is the requirement for equal-dimensional data and latent spaces. However, real-world data often reside on a low-dimensional manifold~\cite{bengio2014}, making a strict equal-dimensional transformation highly restrictive. To address this, injective flows modify the change-of-variables formula for rectangular Jacobians~\cite{kothe2023}. Common strategies include adding noise and denoising to the manifold~\cite{horvat2021}, constraining off-manifold latent variances~\cite{beitler2021,silvestri2023}, or using a two-stage procedure where an autoencoder is trained first, followed by a normalizing flow on its latent space~\cite{brehmer2020,kothari2021,bohm2022a}. Other methods impose strict geometric restrictions, such as conformal~\cite{ross2021} or isometric~\cite{cramer2023} embeddings, which limit expressivity. Rectangular Flows~\cite{caterini2021} introduced iterative, unbiased estimators for the Jacobian log-determinant, but they remain computationally expensive.

\paragraph{Flow Autoencoders.}

Recent work relaxes exact architectural invertibility by using
separately parameterized encoder and decoder networks, trained as
approximate inverses through a reconstruction
penalty~\cite{sorrenson2024,draxler2024,sorrenson2024b}. These models
retain likelihood-based training while permitting flexible,
off-the-shelf architectures. To avoid expensive log-determinant
computations for general networks, existing methods use encoder and
decoder surrogates. Although the two 
surrogates approximate their corresponding log-determinant gradients,
prior work reports stable training primarily when using encoder
surrogate, whereas the decoder surrogate often leads to instability.

Our work provides the first theoretical analysis of the interaction
between the surrogate and reconstruction gradients in flow
autoencoders. We then introduce a conditional loss that dynamically uses both the encoder and decoder surrogates, unlocking the full potential of flow autoencoders and establishing NAE as an effective generative model.

\section{Methodology}

To get insights into the training behaviour of the surrogate losses, we first review the surrogate terms and then derive their relation to the
reconstruction loss.

\subsection{Flow Autoencoders and Approximate Inverses}
\label{sec:approximate_inverses}
Let $E_\theta:\mathbb{R}^D\rightarrow\mathbb{R}^d$ and
$D_\phi:\mathbb{R}^d\rightarrow\mathbb{R}^D$ denote the encoder and
decoder, respectively. Together, they define the autoencoder
\[
f_\psi(x)=D_\phi(E_\theta(x)),
\qquad \psi=(\theta,\phi).
\]
Let $J_\theta$ denote the encoder Jacobian at $x_i$, and
$G_\phi$ the decoder Jacobian at $z_i=E_\theta(x_i)$. When the encoder and decoder are exact inverses of each other and $d=D$, 
the change-of-variables formula gives two equivalent forms,
\begin{equation}
	\begin{aligned}
		\log p_X(x_i)
		&=\log p_Z(z_i)+\log|\det J_\theta| \\
		&=\log p_Z(z_i)-\log|\det G_\phi|.
	\end{aligned}
	\label{nll:og}
\end{equation}

For injective flows ($d<D$), the induced density is defined on the
decoder manifold, and the volume change is determined by the
corresponding Gram matrix~\cite{krantz2008}. The two possible $\operatorname{nll}$ expressions become:
\begin{equation}
	\begin{aligned}
		\nll_E(x_i)
		&=-\log p_Z(z_i)
		-\tfrac12\log\det(J_\theta J_\theta^\top),\\
		\nll_D(x_i)
		&=-\log p_Z(z_i)
		+\tfrac12\log\det(G_\phi^\top G_\phi).
	\end{aligned}
	\label{eq:nll}
\end{equation}
Both expressions reduce to Eq.~\eqref{nll:og} when $d=D$ and the networks are exact inverses.

To relax the exact invertibility constraint, \citet{sorrenson2024} trains the autoencoder by minimizing the negative log-likelihood ($\operatorname{nll}$) under a reconstruction penalty that enforces approximate inversion. The final loss used to train the flow autoencoder is given by:
\begin{equation}
	\mathcal{L}
	=\sum_{i=1}^{n}
	\left[
	\nll(x_i)
	+\beta\|x_i-f_\psi(x_i)\|_2^2
	\right],
	\label{eq:lagrangian}
\end{equation}
where $\beta \geq 0$ is a Lagrange multiplier controlling the reconstruction constraint and will be referred to as \textit{reconstruction weight}.  When $\beta$ is sufficiently large, the encoder and decoder behave as approximate inverses and either $\operatorname{nll}$ term from Eq.~\eqref{eq:nll} can, in principle, drive training.

\subsection{Surrogate Gradient Estimation}
\label{sec:surrogate}

For general encoder and decoder architectures, evaluating the
log-determinants in Eq.~\eqref{eq:nll} is computationally expensive.
Under the approximate-inverse assumption, \citet{sorrenson2024}
approximates their parameter gradients as
\begin{equation}
	\begin{aligned}
		\nabla(
		\tfrac12\log\det(J_\theta J_\theta^\top))
		&\approx
		\tr\!\left((\nabla J_\theta)G_\phi\right),
		\\
		\nabla(
		\tfrac12\log\det(G_\phi^\top G_\phi))
		&\approx
		\tr\!\left(J_\theta(\nabla G_\phi)\right).
		\label{eq:logdet_gradients}
	\end{aligned}
\end{equation}
This estimator was subsequently extended to the full-dimensional
setting ($d=D$) by \citet{draxler2024}.

To calculate the trace, \citet{sorrenson2024} applies the Hutchinson trace estimator in latent space, where the smaller matrix dimension generally reduces the estimator variance. The probe vectors \(v_k\) are chosen as the first \(K\) columns of a randomly sampled orthonormal matrix \(V\in\mathbb{R}^{d\times d}\) and scaled by \(\sqrt{d}\). Consequently, \(v_k^\top v_k=d\) and \(\mathbb{E}[v_kv_k^\top]=I_d\). When \(d=D\), the two trace orderings have comparable variance, and the trace is therefore estimated in data space \citep{draxler2024}.


In practice, the trace is estimated using $K$ vector-Jacobian and Jacobian-vector products along with \texttt{stop\_gradient} operator $\SG(\cdot)$, which can be efficiently calculated in standard autodiff libraries like PyTorch. The corresponding scalar surrogates are given as: 
\begin{align*}
	\mathtt{surr}_E
	&=-\frac{1}{K}\sum_{k=1}^K
	v_k^\top J_\theta\SG(G_\phi v_k),
	\nonumber\\
	\mathtt{surr}_D
	&= \,\,\,\,\, \frac{1}{K}\sum_{k=1}^K
	\SG(v_k^\top J_\theta)G_\phi v_k.
\end{align*}
Backpropagating through these expressions produces the gradient
estimates in Eq.~\eqref{eq:logdet_gradients}. We therefore optimize
\begin{equation}
	\begin{aligned}
		\widetilde{\nll}_E&=-\log p_Z(z_i)+\mathtt{surr}_E, \\
		\widetilde{\nll}_D&=-\log p_Z(z_i)+\mathtt{surr}_D. \label{eq:surr}
	\end{aligned} 
\end{equation}

The first surrogate ($\mathtt{surr}_{E}$) updates the encoder through $J_\theta$, whereas
the second ($\mathtt{surr}_{D}$) updates the decoder through $G_\phi$. We denote them as \textit{Encoder} and \textit{Decoder} surrogates, respectively. 

\subsection{Reconstruction Loss Implicitly Regularizes the Jacobian}

Continuous normalizing flows typically require discrete data, such as pixel-valued images, to be dequantized before density estimation. More generally, local perturbations can be added to the inputs to smooth the empirical data distribution~\cite{kim2020,ho2019}. We show that minimizing the reconstruction loss under such perturbations implicitly regularizes the autoencoder’s Jacobian.

Let $X=x_i+\epsilon$ denote a perturbed observation, where the
centered perturbation satisfies
\[
\mathbb{E}[\epsilon]=0,
\qquad
\mathbb{E}[\epsilon\epsilon^\top]=\Sigma_i.
\]
Here, $\epsilon$ may represent dequantization noise or explicit
local augmentation. It may also model natural local variation,
provided that such variation is approximately centered at $x_i$.  Let
$M_i=G_\phi J_\theta \in \mathbb{R}^{D \times D}$ denote the composite Jacobian of
$f=D_\phi\circ E_\theta$ at $x_i$. For sufficiently small
perturbations, a first-order expansion gives
\begin{equation}
	f(x_i+\epsilon)
	\approx f(x_i)+M_i\epsilon.
	\label{eq:expansion}
\end{equation}

Writing $r_i=x_i-f(x_i)$ and $A_i=I_D-M_i$, the expected local
reconstruction loss becomes
\begin{align}
	\mathcal{R}_i
	&:=\mathbb{E}_\epsilon
	\left[\|X-f(X)\|_2^2\right] \nonumber\\
	&\approx\mathbb{E}_\epsilon
	\left[\|r_i+A_i\epsilon\|_2^2\right] \nonumber\\
	&=\|r_i\|_2^2
	+2r_i^\top A_i\mathbb{E}[\epsilon]
	+\tr(A_i^\top A_i\Sigma_i) \nonumber\\
	&=\|x_i-f(x_i)\|_2^2
	+\|A_i\Sigma_i^{1/2}\|_F^2,
	\label{eq:recon_M_general}
\end{align}
where the cross term vanishes because $\mathbb{E}[\epsilon]=0$.
Thus, reconstruction training encourages $M_i$ to act as the identity
along directions represented by the local covariance $\Sigma_i$.

We follow the setting of \citet{sorrenson2024}, where isotropic noise is added to input samples
during training, such that $\Sigma_i=\sigma^2I_D$. Therefore,
Eq.~\eqref{eq:recon_M_general} reduces to
\begin{equation}
	\mathcal{R}_i
	=\|x_i-f(x_i)\|_2^2
	+\sigma^2\|I_D-M_i\|_F^2.
	\label{eq:recon_M_relation}
\end{equation}
Up to normalization, the overall reconstruction objective is the sum
of these local expectations:
\begin{equation}
	\mathcal{L}_{\mathrm{rec}}
	=\sum_{i=1}^n\mathcal{R}_i.
	\label{eq:recon_decomposition}
\end{equation}

\paragraph{Data-Space to Latent-Space Relation.}

The matrix $M_i$ acts in the data space. Analogously, we define $N_i = J_{\theta} G_{\phi} \in \mathbb{R}^{d \times d}$, which acts in the latent space. For brevity, we omit the sample and parameter subscripts. By the cyclic invariance of the trace,
\begin{equation}
	\tr(M)=\tr(GJ)=\tr(JG)=\tr(N).
	\label{eq:two_spaces}
\end{equation}

Under approximate-inverse assumption, $N$ is close to $I_d$, while $M$ is close to a rank-$d$ orthogonal projector $P_d$ (when $d=D$, $M$ is close to $I_D$):
\[
M=P_d+E_D,
\qquad
N=I_d+E_d.
\]
Since
$\|P_d\|_F^2=\|I_d\|_F^2=d$, defining
$\delta=\|E_D\|_F+\|E_d\|_F$ and applying the Cauchy--Schwarz inequality yields
\begin{equation}
	\left|\|M\|_F^2-\|N\|_F^2\right|
	\leq 2\sqrt{d}\,\delta+\delta^2.
	\label{eq:norm_approx}
\end{equation}
Thus, for small residuals,
$\|M\|_F^2\approx\|N\|_F^2$. Moreover,
\begin{align}
	\|I_D-M\|_F^2
	&=D-2\tr(M)+\|M\|_F^2, \nonumber\\
	\|I_d-N\|_F^2
	&=d-2\tr(N)+\|N\|_F^2.
\end{align}
Combining these identities with Eq.~\eqref{eq:two_spaces} and Eq.~\eqref{eq:norm_approx} yields
\begin{equation}
	\|I_D-M\|_F^2
	\approx\|I_d-N\|_F^2+D-d.
	\label{eq:data_latent_relation}
\end{equation}

When J and G are exact Moore–Penrose inverses, the relation is exact. As $D$ and $d$ are constants, minimizing $\mathcal{L}_{\mathrm{rec}}$~(Eq.~\ref{eq:recon_decomposition}) is approximately equivalent to
minimizing
\begin{equation}
	\begin{aligned}
		\widetilde{\mathcal{L}}_{\mathrm{rec}}
		 \approx \sum_{i=1}^n\bigl[
		\|x_i-f(x_i)\|_2^2\
		+\sigma^2\|I_d-N\|_F^2
		\bigr].
	\end{aligned}
	\label{eq:recon_N_relation}
\end{equation}

\subsection{Conditional Surrogate Loss}
\label{sec:relation_to_surrogate}

Differentiating the squared deviation \(\|I_d-N\|_F^2\) with respect to the model parameters reveals a direct connection to the surrogate losses:
\begin{align}
	\nabla \widetilde{\mathcal{L}}_{\mathrm{rec}}
	&\propto
	\nabla\|I_d-N\|_F^2 \nonumber\\
	&\propto
	\tr\!\left[(N-I_d)^\top\nabla N\right]\\
	\label{eq:recon_gradient}
	& \qquad \text{where} \nabla N=(\nabla J)G+J(\nabla G). \nonumber
\end{align}

We'll use the scaled orthonormal matrix $V$ of size $d \times d$ ($V V^\top = V^\top V = d \cdot I_d$, which was introduced in Sec~\ref{sec:surrogate}) and the cyclic invariance of the trace:
\begin{equation*}
	\begin{aligned}
		\nabla \widetilde{\mathcal{L}}_{\mathrm{rec}}  &\propto \operatorname{tr}\left[(N - I_d)^\top  V V ^\top( \nabla N) V V ^\top\right]\\
		 &\propto \operatorname{tr}\left[V^\top(N - I_d)^\top  V V ^\top( \nabla N) V \right]\\
		 &\propto \operatorname{tr}\left[ \left(V^\top(N - I_d)  V\right)^\top \left(V ^\top( \nabla N) V \right)\right] \\
		 &\propto \sum_{i=1}^d \sum_{j=1}^d \left[v_i^\top (N - I_d) v_j\right] \cdot \left[v_i^\top (\nabla N) v_j\right]. \\
	\end{aligned}
\end{equation*}
Finally, focusing on the term with equal indices ($i=j$),
\begin{align*}
	\nabla \widetilde{\mathcal{L}}_{\mathrm{rec}^{ii}} &\propto \big[v_i^\top (N - I_d) v_i\big] \cdot \left[v_i^\top (\nabla N) v_i\right]\\ 
	&\propto c_{ii} \cdot \big[{v_i^\top (\nabla J)G v_i +v_i^\top J(\nabla G) v_i} \big]\\
	&\propto c_{ii} \cdot \big[{\texttt{\textbf{SG}}(v_i^\top J)G  v_i} + {v_i^\top  J \texttt{\textbf{SG}}(G v_i) } \big].
\end{align*}

\begin{figure}[t]
	\centering
	\includegraphics[width=0.98\columnwidth]{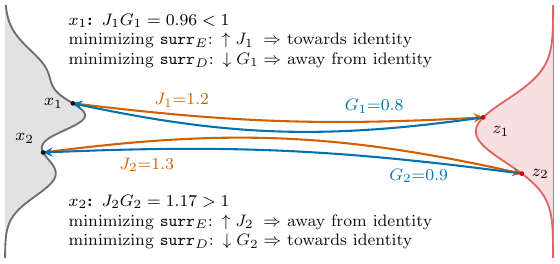} 
	\caption{
		Two points $x_1, x_2$ are mapped from a data space $X$ to a latent space $Z$ via encoder and back via decoder, with scalar Jacobians $J_i, G_i$. For $x_1$, we have $J_1G_1 = 0.96 < 1$ (contractive): the reconstruction gradient is aligned with the $\mathtt{surr}_{E}$, which pushes the composite Jacobian toward identity; but anti-aligned with the $\mathtt{surr}_{D}$, which pushes the composite away from identity. The roles are reversed for $x_2$,~$J_2G_2 = 1.17 > 1$ (expansive). 
	}
	\label{fig:1D_example}
\end{figure}

Crucially, for a given probe vector $v_i$, the scalar prefactor $c_{ii} = \texttt{\textbf{SG}}(v_i^\top (JG - I_d) v_i)$ determines the behaviour of the composite Jacobian in the subspace of the probe vector. When the Jacobian is locally expansive ($c_{ii}>0$):
\begin{equation*}
	\nabla \widetilde{\mathcal{L}}_{\mathrm{rec}^{ii}} \propto  \underbrace{+v_i^\top  J \, \texttt{\textbf{SG}}(G v_i)}_{\textcolor{red}{-{surr}_{E_i}}} \  \underbrace{+ \, \texttt{\textbf{SG}}(v_i^\top J)\, G  v_i}_{\textcolor{blue}{{surr}_{D_i}}}.
\end{equation*}
Similarly, when the Jacobian is locally contractive ($c_{ii}<0$):
\begin{equation*}
	\nabla \widetilde{\mathcal{L}}_{\mathrm{rec}^{ii}} \propto  \underbrace{-v_i^\top  J \, \texttt{\textbf{SG}}(G v_i)}_{\textcolor{blue}{{surr}_{E_i}}} \ \underbrace{-\texttt{\textbf{SG}}(v_i^\top J) \,G v_i}_{\textcolor{red}{-{surr}_{D_i}}}.
\end{equation*}
Within each probe subspace, one surrogate term (\textcolor{blue}{blue}) aligns with the reconstruction-loss gradient, while the other (\textcolor{red}{red}) opposes it. Fig.~\ref{fig:1D_example} illustrates this for a 1D case.

\begin{figure}[b]
	\centering
	\includegraphics[width=0.47\textwidth]{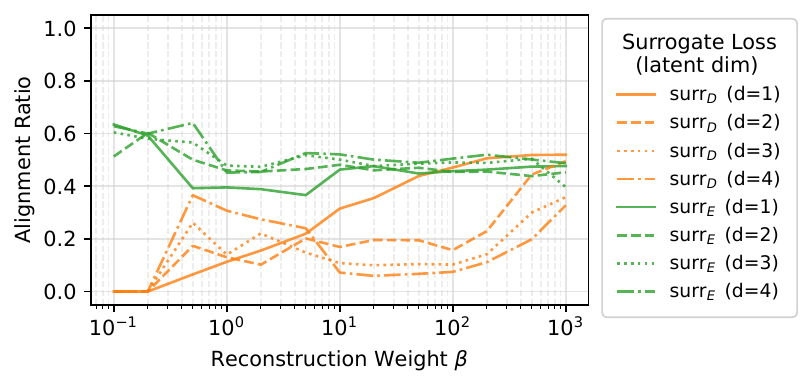} 
	\caption{Alignment ratio (higher is better) averaged over 10 runs for models trained on 4D toy dataset with varying reconstruction weights $\beta$ and latent dim $d$.}
	\label{fig:anti_alignment}
\end{figure}

We investigate this phenomenon further using a 4D toy dataset (Sec.~\ref{sec:toy_problem}). During training, we track the proportion of instances in which the surrogate and reconstruction-loss gradients align. The evolution of alignment throughout training is deferred to Appendix~\ref{app:toy_problem_align}, while Fig.~\ref{fig:anti_alignment} reports the final ratios. Models trained with $\mathtt{surr}_{E}$ maintain moderate alignment, whereas those trained with $\mathtt{surr}_{D}$ exhibit low alignment across most values of $\beta$, indicating that their gradients predominantly oppose the reconstruction-loss gradient. We hypothesize that this opposition causes training instability.

Based on this observation, we propose \textbf{Conditional Loss} that, for each probe, dynamically selects the surrogate term whose gradient is aligned with reconstruction-loss.
Algorithm~\ref{alg:latent_space} provides pseudocode for the conditional loss in bottleneck setting. In full-dimensional setting, the trace is computed in data-space by swapping the JVP and VJP operations. Our method builds upon \citet{sorrenson2024, draxler2024} and introduces minimal computational overhead, specifically Lines 8–11.

\begin{algorithm}[t]
	
	\textbf{Input}:  Batch input $x$, encoder $E_\theta$, decoder $D_\phi$, latent dim $d$, number of Hutchinson probes $K$\\
	\textbf{Output}: Surrogate loss $\mathcal{L}_{\text{cond}}$, encoder output $z$, decoder output $\hat{x}$
	\begin{algorithmic}[1]
		\STATE Let $B = \text{batch size}$
		\STATE Generate Probes $V$ of shape $(K, B, d)$
		\STATE {$ \mathtt{acc}\leftarrow 0$}
		\FOR{$k = 1 \dots K$}
		\STATE $z, J_k \leftarrow \text{VJP}(E_\theta, x, v_k)$ \COMMENT{$J_k = v_k^\top (\nabla E_\theta)$}
		\STATE $\hat{x}, G_k \leftarrow \text{JVP}(D_\phi, z, v_k)$ \COMMENT{$G_k = (\nabla D_\phi) v_k$}
		\STATE $\mathtt{surr}_E \leftarrow -J_k \cdot \SG(G_k)$ \COMMENT{detach $G_k$}
		\STATE $\mathtt{surr}_D \leftarrow \SG(J_k) \cdot G_k$ \COMMENT{detach $J_k$}
		\STATE $s_k \leftarrow \|v_k\|^2$
		\STATE $\mathtt{mask} \leftarrow \mathbf{1}[\mathtt{surr}_D > s_k]$ \COMMENT{binary indicator}
		\STATE $\ell_k \leftarrow \SG(\mathtt{mask}) \cdot \mathtt{surr}_D + (1 - \SG(\mathtt{mask})) \cdot \mathtt{surr}_E$
		\STATE $\mathtt{acc} \leftarrow \mathtt{acc} + \ell_k$
		\ENDFOR
		\STATE $\mathcal{L}_{\text{cond}} \leftarrow {\mathtt{acc}}/{K}  $ \COMMENT{Surrogate loss to be minimized}
		\RETURN $\mathcal{L}_{\text{cond}}, z, \hat{x}$
	\end{algorithmic}
	\caption{Conditional Loss}
	\label{alg:latent_space}
\end{algorithm}

\section{Experiments}

We begin this section by evaluating our method on a synthetic toy dataset. We then evaluate the method first in the full-dimensional setting, and subsequently in the bottleneck setting. We use a single probe ($K=1$) for Hutchinson trace estimation in all our experiments unless stated otherwise.

\begin{figure*}[t]
	\centering
	\includegraphics[width=0.98\textwidth]{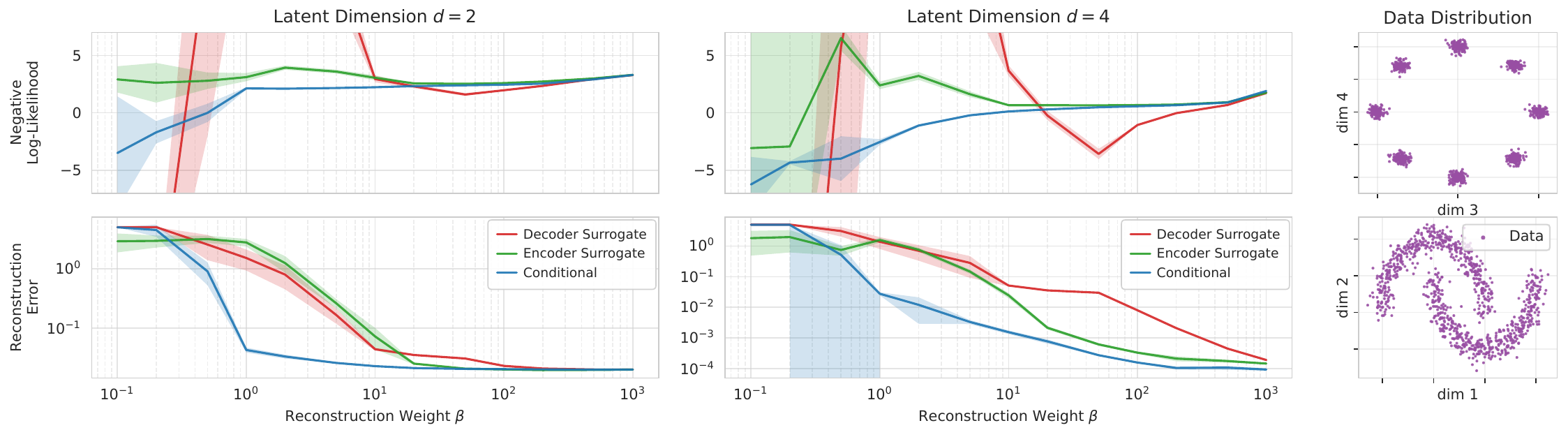} 
	\caption{4D Toy Dataset: \textit{Left} and \textit{middle} columns show test results for latent dimensions $d=2$ and $d=4$ respectively. The top row shows $\operatorname{nll}$ ({lower is better}), while the bottom row shows Reconstruction Error in log-scale ({lower is better}). Shaded regions indicate $\pm 1$ standard deviation over 10 runs. \textit{Right} column illustrates the toy data distribution; dim 1 and 2 corresponds to Two-moons and dim 3 and 4 corresponds to 8-GMM (eight Gaussian clusters evenly spaced along a circle).}
	\label{fig:toy_problem}
\end{figure*}

\subsection{4D Toy Dataset}
\label{sec:toy_problem}

We construct a 4D synthetic dataset by concatenating samples from two 2D distributions: two moons and an eight-component GMM (data distribution shown in Fig.~\ref{fig:toy_problem}). We evaluate all latent dimensions $d\in\{1,2,3,4\}$.

We train a simple MLP autoencoder using three methods: the \textbf{Decoder Surrogate}, \textbf{Encoder Surrogate}, and proposed \textbf{Conditional} loss. Each method is evaluated over a broad range of reconstruction weights $\beta$.

Fig.~\ref{fig:toy_problem} presents the results for $d=2$ and $d=4$; complete results for $d=1$ and $d=3$, along with training details, are provided in Appendix~\ref{app:toy_problem}. We make the following observations:

\begin{itemize}
	\item When training models across different $\beta$ values, each method reaches the approximate-inverse regime at a different $\beta$ value. For $d<D$, this transition value can be estimated using the elbow method heuristic on the reconstruction loss versus $\beta$ graph.
	\item Training a model with a $\beta$ higher than this transition value degrades the $\operatorname{nll}$ while yielding marginal improvements in reconstruction error.
	
	\item Since the proposed \textit{Conditional} loss is aligned with the reconstruction objective by construction, compared to the other methods:
	\begin{itemize}
		\item it requires a significantly smaller $\beta$ to reach the approximate-inverse regime;
		\item it achieves lower reconstruction error and $\operatorname{nll}$ within this regime.
	\end{itemize}
\end{itemize}

\subsection{Full-Dimensional Setting}

In this section, we apply Normalizing Autoencoders (NAEs) to molecular generation. The objective is to model the joint probability distribution over atomic coordinates $x = (x_1, \dots, x_N) \in \mathbb{R}^{N \times n}$ for the DW4, LJ13, and LJ55 datasets, and over both coordinates and atomic properties (atom type and charge) for the QM9 dataset. 

\paragraph{Equivariant Models}
Molecular systems exhibit Euclidean symmetry such that their physical properties are invariant to rotations and translations. Prior work \cite{kohler2020, hoogeboom2022a} has shown that explicitly incorporating this inductive bias during training is more effective than relying on data augmentation. Therefore, we employ an $E(n)$-Equivariant GNN \cite{satorras2022a} and adopt the exact model architecture, hyperparameters, and evaluation protocols from \citet{draxler2024} without tuning. Implementation details are provided in Appendix~\ref{app:en-gnn}.

\paragraph{Boltzmann Generator}
We evaluate our model as a Boltzmann generator, which models the probability of a specific atomic configuration. The probability of a configuration depends on its energy via the Boltzmann distribution:
\begin{equation}
	q(x) \propto e^{-\beta_t u(x)}, \label{eq:boltzman_prob}
\end{equation}
where $\beta_{t}$ is the inverse temperature of the system, and $u(x)$ is the energy function, which takes atomic positions $x~=~(x_1, \dots, x_N)$ as input.

We benchmark our method on \textbf{DW4} (four particles in 2D space interacting via a double-well potential), \textbf{LJ13} and \textbf{LJ55} (13 and 55 particles, respectively, in 3D space interacting via a Lennard-Jones potential). Further details of the DW and LJ potentials are provided in Appendix~\ref{app:bg}. In all tasks, the total energy of the system is the sum of pairwise interactions $u(x) = \sum_{i,j} v(x_i, x_j)$.

\begin{table}[t]
	\small
	\centering
	\setlength{\tabcolsep}{1mm}
	\begin{tabular}{l|rr|rr|rr} 
		\hline
		& \multicolumn{2}{c|}{\textbf{DW4}} & \multicolumn{2}{c|}{\textbf{LJ13}} & \multicolumn{2}{c}{\textbf{LJ55}} \\
		{Method}& \multicolumn{1}{c}{$\operatorname{nll}$} & \multicolumn{1}{c|}{ST} & \multicolumn{1}{c}{$\operatorname{nll}$} & \multicolumn{1}{c|}{ST} & \multicolumn{1}{c}{$\operatorname{nll}$} & \multicolumn{1}{c}{ST} \\
		\hline
		\rowcolor{rowOdd} {$E$-NF} & 1.72$\pm$.01 & \textbf{.024} & -16.28$\pm$.04 & .27 & \multicolumn{2}{c}{OOM} \\
		\rowcolor{rowEven} {OT-FM} & 1.70$\pm$.02 & .034 & -16.54$\pm$.03 & .77 & -88.45$\pm$.04 & 40.0 \\
		\rowcolor{rowOdd} {$E$-OT-FM} & 1.68$\pm$.01 & .033 & -16.70$\pm$.12 & .72 & -89.27$\pm$.04 & 40.0 \\
		\rowcolor{rowEven} {$E$-FFF} & 1.68$\pm$.01 & .026 & -17.09$\pm$.16 & \textbf{.11} & -88.72$\pm$.16 & \textbf{2.1} \\
		\rowcolor{rowOdd} {{$E$-NAE}} & \textbf{1.39$\pm$.01} & .026 & \textbf{-17.95$\pm$.16} & \textbf{.11} & \textbf{-92.32$\pm$.09} & \textbf{2.1} \\
		\hline
	\end{tabular}
	\caption{Results on DW4, LJ13, LJ55 datasets: Average values with standard deviation over 3 runs for $\operatorname{nll}$ (lower is better) and ST (sampling time in ms, lower is better) are reported. \textbf{Bold} indicates the best performance. OOM:out-of-memory. Baseline results are taken from FFF.}
	\label{tab:mol_results}
\end{table}

For evaluation, we report the $\operatorname{nll}$ (under the Boltzmann distribution in Eq.~\ref{eq:boltzman_prob}) and the time required to generate samples. 
In Table~\ref{tab:mol_results}, we compare against Equivariant ODE normalizing flow ({$E$-NF}) \cite{satorras2022a}; ODEs trained via optimal transport flow matching, both with equivariance ($E$-OT-FM) and without (OT-FM) \cite{klein2023}; and Equivariant Free-Form Flow ($E$-FFF), which employs the Encoder surrogate loss \cite{draxler2024}. 

Our method achieves lowest negative log-likelihood on all three datasets. Since NAE and FFF differ solely in their training objective while sharing the same architecture, their inference procedures are functionally equivalent. We therefore use the sampling times reported for FFF as direct proxies for those of NAE.

\paragraph{QM9}

We also evaluate NAE on the QM9 dataset \cite{ramakrishnan2014}, which contains molecules with variable numbers of atoms, with at most 29 atoms. The task requires the model to jointly model atomic coordinates and properties $h_i$ (categorical atom type and ordinal charge).   The network treats the atomic properties $h_i$ as invariant under Euclidean transformations.

In Table~\ref{tab:molecule-qm9}, we compare against Equivariant Diffusion ($E$-DM) \cite{hoogeboom2022a}; Equivariant ODE normalizing flow  ({$E$-NF}) \cite{satorras2022a}; and Equivariant Free-Form Flow ($E$-FFF) \cite{draxler2024}. 
Our method achieves the best $\operatorname{nll}$ and Stable Sampling Time, while being competitive on molecular stability among non-diffusion methods.

\textit{Stable Sampling Time} denotes the average time required to generate a stable molecule, including the time spent generating unstable samples that are subsequently discarded. While $E$-DM yields higher molecular stability, it does so at the cost of sampling time which is orders of magnitude slower. This highlights a key advantage of NAE over diffusion-based approaches, which are often preferred for generation tasks but suffer from iterative sampling overhead.

\begin{table}[t]
	\small
	\centering
	\begin{tabular}{l|r|r|rr} 
		\hline
		& & \multicolumn{1}{c|}{Molecular} & \multicolumn{2}{c}{Sampling Time } \\
		Method& $\operatorname{nll}$ ($\downarrow$) & Stability ($\uparrow$)   & Raw($\downarrow$) & Stable ($\downarrow$)\\ \hline 
		\rowcolor{rowOdd} $E$-DM  & {-110.7} & \textbf{82.0\%} & 1580.8 & 1970.6 \\
		\rowcolor{rowEven} $E$-NF  & -59.7 & 4.9\% & 13.9 & {309.5} \\
		\rowcolor{rowOdd} $E$-FFF & -76.2 & 8.7\% & \textbf{0.6} & {8.1} \\
		\rowcolor{rowEven}$E$-NAE &\textbf{ -120.8}  & 9.3\% & \textbf{0.6} & \textbf{7.5} 
		\\ \hline 
	\end{tabular}
	\caption{%
		Results on QM9 molecule Dataset. Validation $\operatorname{nll}$ (lower is better), molecular stability of generated samples (higher is better) and ST (sampling time in ms, lower is better) are reported. \textbf{Bold} indicates the best performance. Baseline results are taken from FFF.}
	\label{tab:molecule-qm9}
\end{table}


\subsection{Bottleneck Setting}

In this section, we first evaluate the generative performance of NAE on tabular datasets. We then assess our model on the Pythae image generation benchmark. For both tasks, we adopt the model architectures, hyperparameters, and evaluation protocols from \citet{sorrenson2024}. Finally, we compare our method against state-of-the-art injective flows. Implementation details are provided in Appendices~\ref{app:tabular_exp} and~\ref{app:pythae_exp}.


\begin{table}[b]
	\setlength{\tabcolsep}{1mm}
	\small
	\centering{
		\begin{tabular}{l|c|c|c|c}
			\hline
			Method & \textbf{Power}  & \textbf{Gas}  & \textbf{HEPMASS}  & \textbf{MiniBooNE} \\
			\hline
			\rowcolor{rowOdd} M-Flow & 0.258$\pm$.045 & 0.219$\pm$.016 & 0.741$\pm$.052 & 1.650$\pm$.105 \\
			\rowcolor{rowEven}RF  & 0.083$\pm$.015 & \textbf{0.110$\pm$.021} & 0.779$\pm$.191 & 1.001$\pm$.051 \\
			\rowcolor{rowOdd} CMF  & 0.053$\pm$.005 & 0.373$\pm$.065 & 0.574$\pm$.044 & 1.508$\pm$.082 \\
			\rowcolor{rowEven} FIF  & \underline{{0.041$\pm$.007}} & 0.281$\pm$.031 & \underline{{0.541$\pm$.034}} & \underline{{0.598$\pm$.024}} \\
			\rowcolor{rowOdd} NAE & \textbf{0.013$\pm$.005} & \underline{{0.216$\pm$.013}} & \textbf{0.485$\pm$.025} & \textbf{0.583$\pm$.027} \\
			\hline
		\end{tabular}
	}
	\caption{%
		Results on tabular datasets. Average FID scores (lower is better) and standard deviations across 5 runs are reported. \textbf{Bold} indicates the best performance, while \underline{{underlined}} indicates the second-best. Baseline results are taken from their respective papers (except for M-flow, where the results are taken from CMF). 
	}
	\label{tab:tabular_results}
\end{table}

\subsubsection{Tabular Datasets}
\label{sec:tabular_datasets}
We evaluate on four tabular datasets from \citet{papamakarios2017a}. We evaluate generative quality using the \textit{FID-like} metric proposed by \citet{caterini2021} and adopt model settings and hyperparameters from \citet{sorrenson2024}. 

In Table~\ref{tab:tabular_results}, we compare against   M-Flow \cite{brehmer2020}; Rectangular Flows (RF) \cite{caterini2021};
Canonical Manifold Flows (CMF) \cite{flouris2023}; and Free-Form
Injective Flows (FIF) \cite{sorrenson2024}. Our method achieves the best performance on three of the four datasets, achieving the lowest score on Power, HEPMASS, and MiniBooNE, while ranking second on Gas.

\subsubsection{Pythae Benchmark}
Following \citet{sorrenson2024}, we benchmark NAE on the Pythae Benchmark \cite{chadebec2022}, which compares various autoencoder methods across two architectures and three datasets. All models are trained for the same number of iterations and evaluated using FID (Fréchet Inception Distance) and IS (Inception Score) under two latent prior distributions: a standard normal ($\mathcal{N}$) and a 10-component Gaussian mixture model (GMM) fit to the encoded training data.
We report CelebA FID scores in Table~\ref{tab:pythae_celeba}, our method outperforms all baselines in three of the four settings and remains competitive in the fourth. Detailed results are deferred to Appendix~\ref{app:pythae_results}.

\begin{table}[t]
	\centering
	\small
	\setlength{\tabcolsep}{1mm}
	\begin{tabular}{l|rr|rr}
		
		\hline
		\multirow{3}{*}{Model} & \multicolumn{2}{c|}{ConvNet} & \multicolumn{2}{c}{ResNet} \\
		& \multicolumn{2}{c|}{(\textbf{33.5M} Params)}& \multicolumn{2}{c}{(\textbf{1.6M} Params)} \\
		& $\mathcal{N}$  & GMM  & $\mathcal{N}$  & GMM  \\
		\hline
		\rowcolor{rowOdd}VAE          & \underline{54.8} & 52.4 & 66.6 & 63.0 \\
		\rowcolor{rowEven}IWAE     & 55.7 & 52.7 & 67.6 & 64.1 \\
		\rowcolor{rowOdd}VAE-lin NF  & 56.5 & 53.3 & 67.1 & 62.8 \\
		\rowcolor{rowEven}VAE-IAF & 55.4 & 53.6 & 66.2 & 62.7 \\
		\rowcolor{rowOdd}$\beta$-(TC) VAE  & 55.7 & 51.7 & 65.9 & 59.3 \\
		\rowcolor{rowEven}FactorVAE  & \textbf{53.8} & 52.4 & 66.4 & 63.3 \\
		\rowcolor{rowOdd}InfoVAE     & 55.5 & 52.7 & 66.4 & 62.3 \\
		\rowcolor{rowEven}AAE & 59.9 & 53.9 & {64.8} & 58.7 \\
		\rowcolor{rowOdd}MSSSIM-VAE  & 124.3 & 124.3 & 119.0 & 119.2 \\
		\rowcolor{rowEven}Vanilla AE                          & 327.7 & 55.4 & 275.0 & {57.4} \\
		\rowcolor{rowOdd}WAE  & 64.6 & 51.7 & 67.1 & 57.7 \\
		\rowcolor{rowEven}VQVAE   & 306.9 & {51.6} & 140.3 & 57.9 \\
		\rowcolor{rowOdd}RAE    & 86.1 & 52.5 & 168.7 & 58.3 \\
		\rowcolor{rowEven} {FIF}                 & 56.9 & \underline{47.3} & \underline{62.3} & \underline{55.0} \\
		\rowcolor{rowOdd}{NAE}    & 57.6  & \textbf{36.9} & \textbf{60.4}  & \textbf{53.4}  \\
		\hline
	\end{tabular}
	\caption{{Pythae benchmark results on CelebA:} FID scores (lower is better). \textbf{Bold} indicates the best performance, while \underline{{underlined}} indicates the second-best. Results taken from FIF.}
	\label{tab:pythae_celeba}
\end{table}

\begin{figure*}[t]
	\centering
	\includegraphics[width=0.99\textwidth]{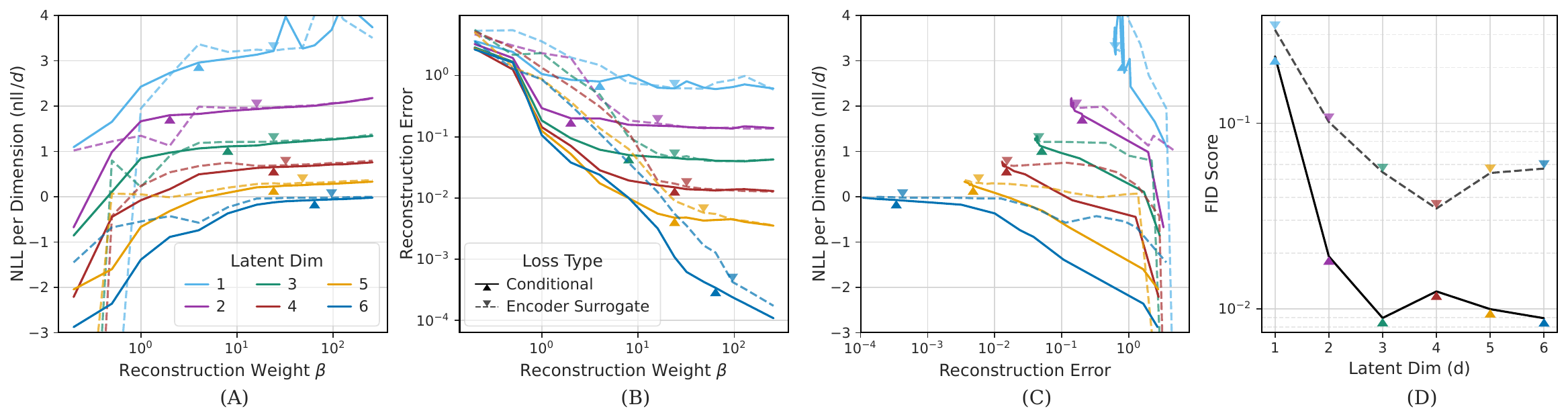} 
	\caption{Ablation study on the Power dataset. Legend is shared across panels. Line Style denote loss type (solid for Conditional and dashed for Encoder Surrogate). Colours denote the latent dimension ($d$). Markers ($\blacktriangle$ for Conditional and $\blacktriangledown$ for Encoder Surrogate) indicate the $\beta$ value that achieves the best FID-like metric for each configuration.
		}
	\label{fig:tabular_ablation}
\end{figure*}

\paragraph{Higher-order derivatives.} 

The models proposed in Pythae employ $\operatorname{ReLU}$ activations. Because the surrogate loss involves second-order derivatives, and $\operatorname{ReLU}''(x)=0$ almost everywhere (with autodiff implementations typically assigning it as zero at the origin), backpropagation through VJP/JVP operations discards all curvature information contributed by the activation functions. To address this limitation, we replace $\operatorname{ReLU}$ with $\operatorname{SiLU}$, whose second derivative is nonzero, continuous, and smooth. In our experiments, this modification further improves the model performance.
Owing to space constraints, we defer the corresponding results to Appendices~\ref{app:tabular_exp} and~\ref{app:pythae_results}.

%


\subsubsection{Injective Models}

Finally, we compare our approach with three injective models on CelebA dataset: DNF \cite{horvat2021}; Trumpets \cite{kothari2021}; and FIF \cite{sorrenson2024}. After observing overfitting with the architecture proposed in FIF, we reduced the model capacity rather than modifying the training pipeline. To establish the comparative computational budget on our hardware, we reproduced the reported FIF performance in 6.5 hours and allocated the same wall-clock time to our method. For evaluation, a 10-component GMM is also fit to the encoded training data.A GMM prior can be beneficial when the learned latent distribution contains disconnected, multimodal, or otherwise non-Gaussian structure. A substantial improvement when sampling from the GMM rather than \(\mathcal{N}\) suggests that its additional flexibility captures latent structure missed by the standard Gaussian. Conversely, a small improvement suggests that the learned latent representation is more regular and better aligned with the assumed Gaussian prior.

\begin{table}[t]
	\centering
	\small
	\setlength{\tabcolsep}{1mm}
	\begin{tabular}{lc|cc|cc|c}
		\hline
		\multirow{2}{*}{Model} & \# Params & \multicolumn{2}{c|}{$\mathcal{N}$ sampler} & \multicolumn{2}{c|}{GMM sampler} & $\Delta$ \\
		& (M) & FID $\downarrow$  & IS $\uparrow$  & FID $\downarrow$ & IS $\uparrow$ & FID  \\
		\hline
		\rowcolor{rowOdd}DNF & 39.4 & 55.6 $\pm$ 0.59 & 1.9 & 52.7 $\pm$ 0.33 & 2.0 & 2.9 \\
		\rowcolor{rowEven}Trumpet & 19.1 & 56.2 $\pm$ 1.39 & 1.8 & 47.7 $\pm$ 2.24 & 1.9 & 8.5 \\
		\rowcolor{rowOdd}FIF & 34.3 & 47.3 $\pm$ 1.39 & 1.7 & \textbf{37.4 $\pm$ 1.35} & 2.0 & 9.9 \\
		\rowcolor{rowEven}NAE & 9.9 & \textbf{45.7 $\pm$ 0.43} & 2.1 & 41.2 $\pm$ 0.54 & 2.0 & 4.5 \\
		\hline
	\end{tabular}
	\caption{Results for injective flows on CelebA. FID (lower is better) and IS (higher is better) are reported averaged over 3 runs with standard deviation. Number of parameters (in millions) of the model are also reported. The last column shows the absolute difference in FID between the samplers. Baseline results are taken from FIF.}
	\label{tab:injective-comparison}
\end{table}

In Table~\ref{tab:injective-comparison}, the strong performance of NAE with \(\mathcal{N}\), together with the small FID gap between the two priors, therefore suggests that it learns a coherent latent representation that can be sampled effectively using a standard Gaussian. However, this small gap alone does not establish that the learned latent distribution is Gaussian. Further discussion is provided in Appendix~\ref{app:pythae_results}.

\subsection{Ablation Study on Tabular Data}

We conduct a controlled ablation study on real-world Power dataset \cite{papamakarios2017a} and compare our method with Encoder Surrogate loss across a range of reconstruction weights $\beta$ and latent dimensions $d$. Implementation details are provided in Appendix~\ref{app:ablation_exp}.

For each configuration, we report the mean over three independent runs of the negative log-likelihood per latent dimension ($\operatorname{nll}/d$), reconstruction error, and FID. 

Fig.~\ref{fig:tabular_ablation}(A,B) show that larger latent dimensions generally require larger values of $\beta$ to reach the approximate-inverse regime. We hypothesize that increasing $d$ alters the relative scales of the likelihood and reconstruction terms, thereby requiring a larger reconstruction weight to balance the joint objective and ensure approximate inversion.


Fig.~\ref{fig:tabular_ablation}(C) further illustrates the trade-off between reconstruction error and $\operatorname{nll}$ as $\beta$ varies: increasing $\beta$ reduces the reconstruction error but increases $\operatorname{nll}$. Since both quantities are minimized, configurations near the lower-left region of the plot are preferable. 

Finally, Fig.~\ref{fig:tabular_ablation}(D) compares generative quality across different latent dimensions. The FID score decreases as \(d\) increases, reaches a minimum at an intermediate dimension, and then increases slightly. One possible explanation is that generative performance is optimized when the latent dimension coincides with the intrinsic dimensionality of the data manifold. A more thorough theoretical investigation of this phenomenon is left for future work.

\section{Limitations and Conclusions}

\paragraph{Limitations.}
Our analysis assumes locally smooth data densities and may therefore be less applicable when this assumption does not hold. Flow autoencoders also introduce an additional hyperparameter, the reconstruction weight \(\beta\), whose appropriate scale may depend on the latent dimension and other model characteristics. For \(d<D\), we recommend first training solely with the reconstruction loss to estimate the minimum attainable reconstruction error, then choosing \(\beta\) such that the training reconstruction error remains within a prescribed tolerance of this minimum.

\paragraph{Conclusions.}
This work presents a theoretical and empirical study of flow autoencoders. We establish that both encoder and decoder surrogate-loss terms are necessary for effective training. We introduce Normalizing AutoEncoder (NAE), which employs a novel conditional loss that ensures alignment of the surrogate loss with the reconstruction objective. We demonstrate NAE's applicability for generative modeling across multiple domains. Our findings also open promising directions for future research, including the influence of activation functions on losses involving second-order derivatives, the extension of NAE to larger-scale datasets, and the exploration of more expressive encoder-decoder architectures for high-dimensional generative modeling.

\appendix
\part*{Appendix}
\section{Compute Details}
All experiments were conducted on a server equipped with 7\(\times\) \textit{NVIDIA RTX A5000} GPUs, 1\(\times\) \textit{NVIDIA RTX PRO 6000} GPU, and two \textit{AMD EPYC 7453} CPU processors.

\section{Toy Problem}
\label{app:toy_problem}
\subsection{Experimental Setup}
\label{app:toy_problem_exp}

All experiments used an MLP-based autoencoder architecture. The encoder and decoder each consisted of a five-layer MLP with a hidden dimension of 256 and SiLU (Swish) activations applied after each hidden layer. Neither network used skip connections. The weights were initialized using Xavier uniform initialization with a gain of $\sqrt{2}$. The input dimension was fixed at 4, while the latent dimension was specified separately for each experiment.

The dataset consisted of four-dimensional samples formed by concatenating two independent two-dimensional sources: (i) eight Gaussian clusters uniformly arranged on a circle of radius 2.0, each with a standard deviation of 0.1, and (ii) two interleaving half-circles generated using the scikit-learn \texttt{make\_moons} function with a noise level of 0.1. A test set of 1,000 samples was generated using the same procedure.

Training was performed for 30,001 iterations with a batch size of 256. We used the Adam optimizer with an initial learning rate of $10^{-3}$ and a cosine annealing schedule spanning the full training duration. Gradients were clipped to a maximum norm of 1.0. The reconstruction loss was defined as the mean squared error (MSE) between the inputs and their reconstructions. A $d$-dimensional standard Gaussian distribution was used as the latent prior. The trace was estimated using a single Hutchinson probe.

For each method and $\beta$ value, we conducted 10 independent runs. Random seeds were matched across methods for each run. Each model was trained on a single CPU core and required approximately 20 minutes.

\begin{figure}[t]
	\centering
	\includegraphics[width=0.43\textwidth]{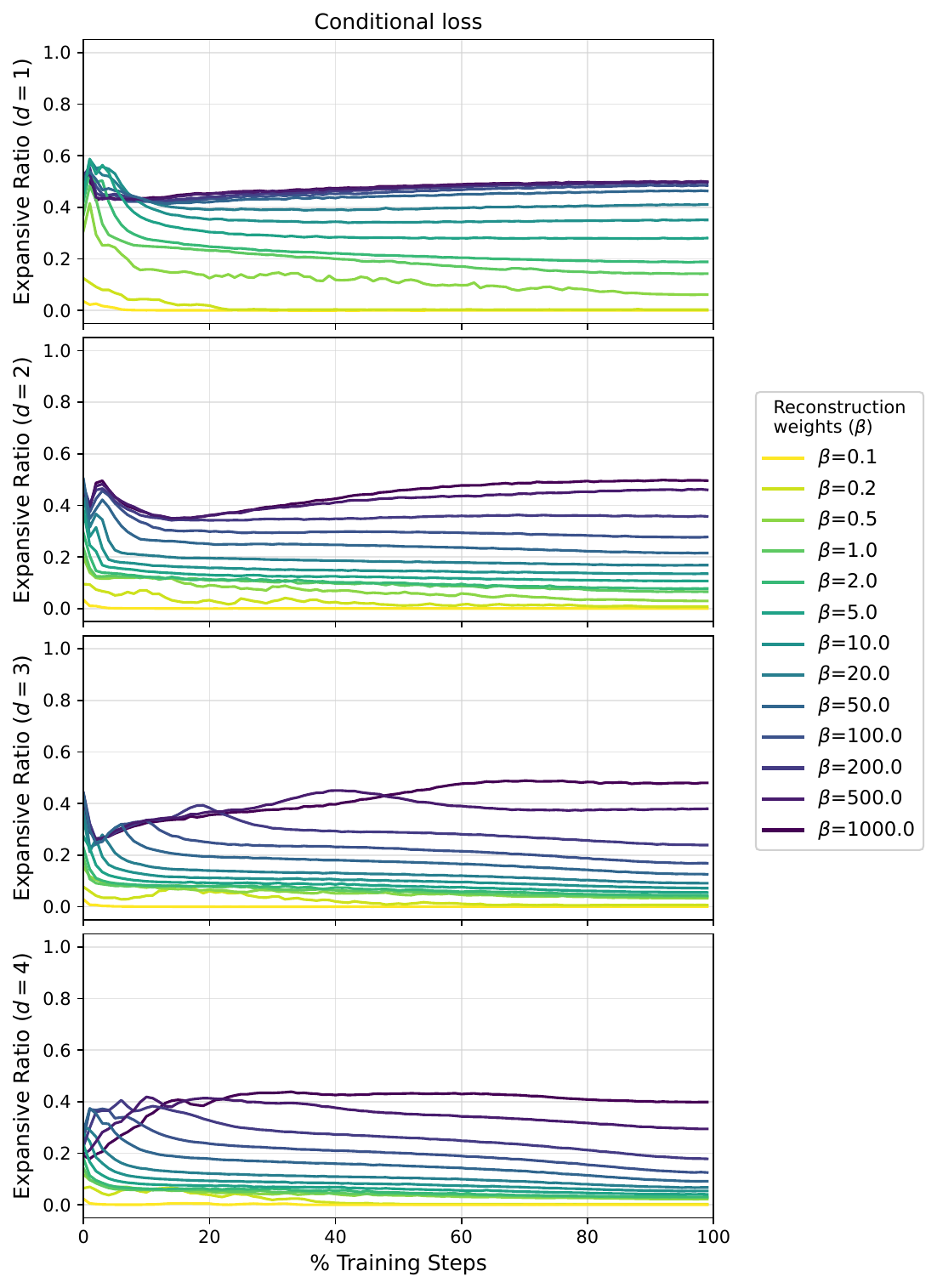} 
	\caption{Expansive Ratio averaged over 10 runs for models trained using Conditional loss on the 4D toy dataset with varying reconstruction weights $\beta$ and latent dim $d$.}
	\label{fig:cond_alignment}
\end{figure}

\subsection{Alignment Ratio}
\label{app:toy_problem_align}

\begin{figure*}[t]
	\centering
	\includegraphics[width=0.85\textwidth]{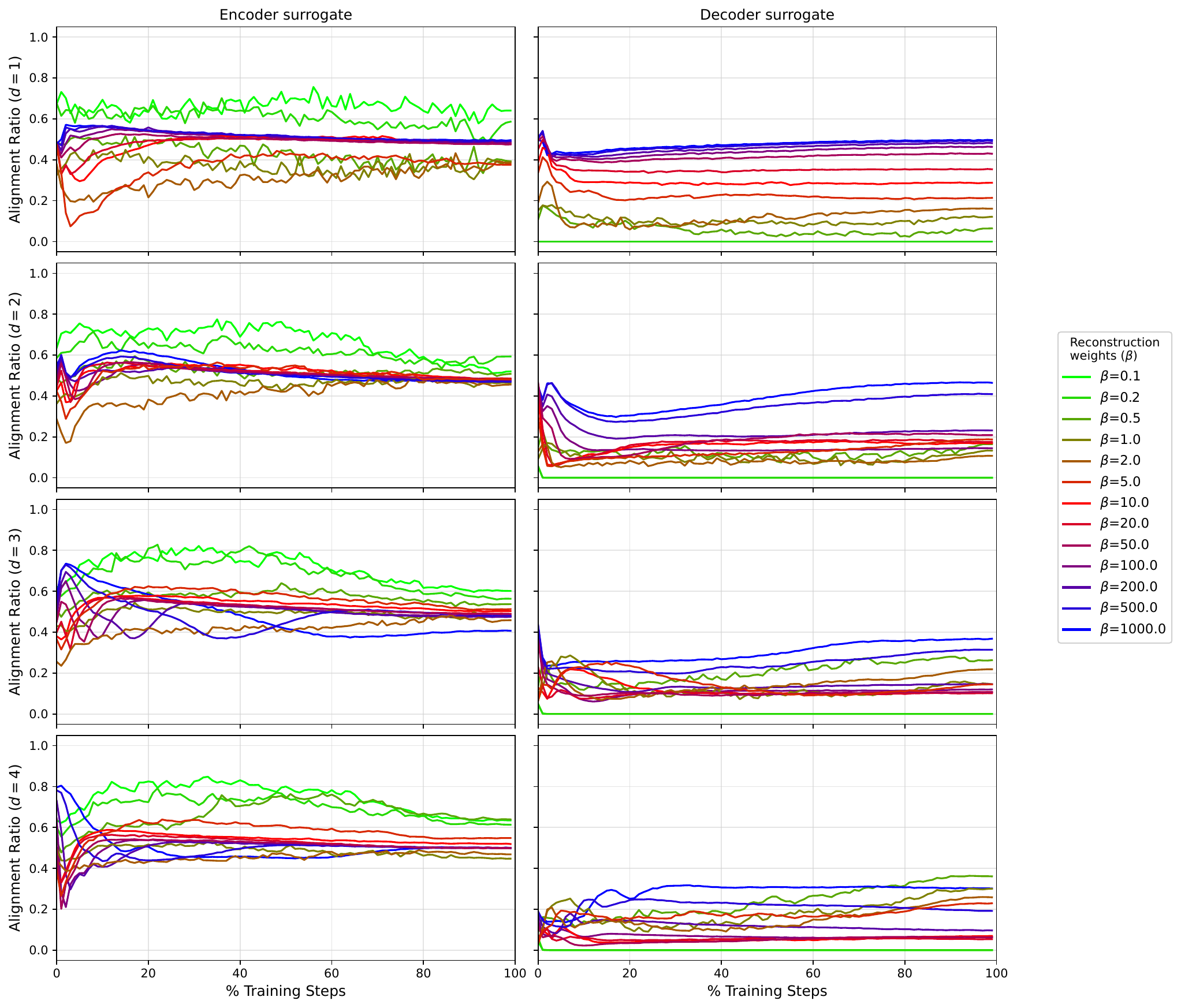} 
	\caption{Alignment Ratios for Encoder and Decoder Surrogates. Alignment ratio (higher is better) averaged over 10 runs for models trained on 4D toy dataset with varying reconstruction weights $\beta$. Left and right columns represent Encoder and Decoder Surrogate losses and the rows represent the latent dim $d$.}
	\label{app:anti_alignment}
\end{figure*}
During training, we tracked the expansive ratio, defined as the proportion of instances in which the Jacobians are locally expansive within the probe subspace. Fig. \ref{fig:cond_alignment} shows how this ratio evolves over the course of training with the Conditional loss. Across all latent dimensions \(d\), the proportion of expansive Jacobians increases with the reconstruction weight \(\beta\). Because a sufficiently large \(\beta\) is required to reach the approximate-inverse regime, a substantial proportion of Jacobians are expected to be expansive once this regime is reached.

For training with the Encoder and Decoder surrogates, we can compute the proportion of updates for which each surrogate is aligned with the reconstruction loss. Fig. \ref{app:anti_alignment} shows how these alignment ratios evolve over the course of training.

For the Decoder surrogate, the alignment ratio increases with \(\beta\) across all latent dimensions but rarely exceeds \(40\%\) when \(d>1\). This limited alignment with the reconstruction objective explains the instability observed when training with the Decoder surrogate alone.

In contrast, the Encoder surrogate maintains an alignment ratio of \(40\text{--}60\%\) across all values of \(\beta\) and \(d\), resulting in consistently more stable training than with the Decoder surrogate.	

\subsection{Extended Results}
\label{app:toy_example_other_results}

\begin{figure*}[t]
	\centering
	\includegraphics[width=0.95\textwidth]{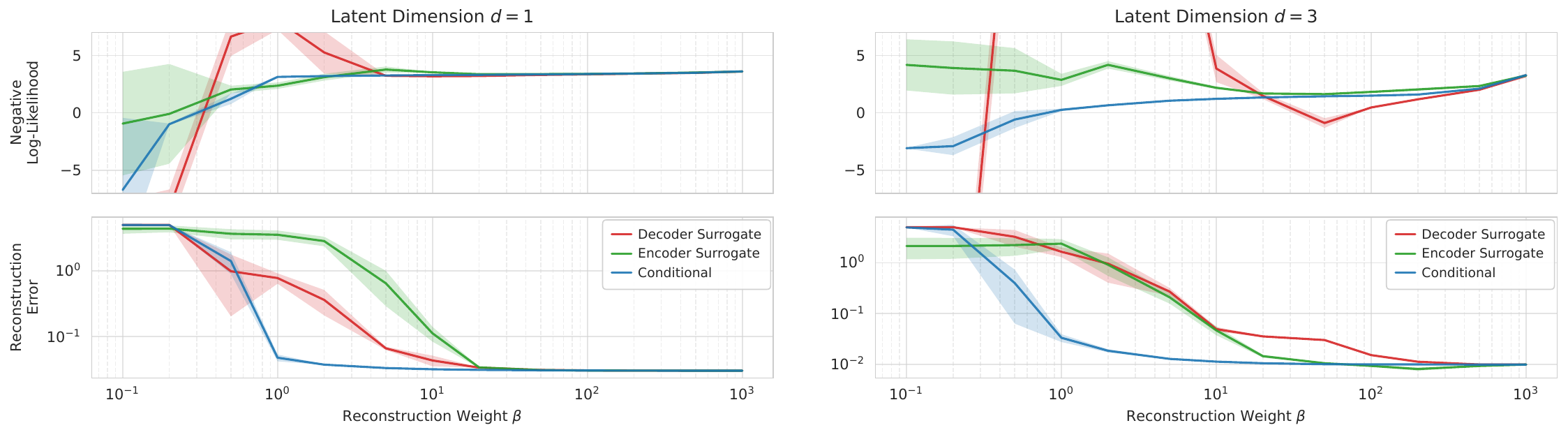} 
	\caption{{4D Toy Dataset}: {Left} and {right} columns show test results for models trained on latent dimensions $d=1$ and $d=3$, respectively. The top row shows Negative Log-Likelihood ($\operatorname{nll}$) ({lower is better}), while the bottom row shows Reconstruction Loss in log-scale ({lower is better}). Shaded regions indicate $\pm 1$ standard deviation over 10 runs.}
	\label{fig:toy_problem_2}
\end{figure*}

Figure \ref{fig:toy_problem_2} presents the results for \(d=1,3\) using both \(\operatorname{nll}\) and reconstruction error. These results are consistent with our observations for \(d=2,4\): each method enters the approximate-inverse regime at a different value of \(\beta\), with the Conditional loss reaching this regime at a significantly lower \(\beta\) than the other methods. Once in this regime, the Conditional loss consistently achieves superior reconstruction and \(\operatorname{nll}\) performance.

We also observe that for $d=1$, the Decoder surrogate performs better than the Encoder surrogate. We hypothesize that this occurs when the latent dimension is smaller than the intrinsic data dimensionality. A more thorough theoretical investigation of this phenomenon is left for future work. 

\begin{table*}[t]
	\centering
	\small
	\begin{tabular}{l|c|c|c|c}
		\hline
		\rowcolor{highlight} & \textbf{DW4} & \textbf{LJ13} & \textbf{LJ55} & \textbf{QM9} \\ \hline
		\rowcolor{rowOdd}Layer count & 20 & 8 & 10 & 16 \\
		\rowcolor{rowEven}Reconstruction weight \(\beta\) & 10 & 200 & 500 & 2000 \\
		\rowcolor{rowOdd}Learning rate & 0.001 & 0.001 & 0.001 & 0.0001 \\
		\rowcolor{rowEven}Optimizer & Adam & Adam & Adam & Adam \\
		\rowcolor{rowOdd}Adam Betas & \([0.9, 0.99]\) & \([0.9, 0.99]\) & \([0.9, 0.99]\) & \([0.9, 0.99]\) \\
		\rowcolor{rowEven}Learning rate scheduler & One cycle & -- & -- / Exponential \(\gamma=0.98\) & -- / Exponential \(\gamma=0.99\) \\
		\rowcolor{rowOdd}Gradient clip & 1 & 1 & 0.1 & 1 \\
		\rowcolor{rowEven}Batch size & 256 & 256 & 48 & 48 \\
		\rowcolor{rowOdd}Epochs & 50 & 400 & 145 / 215 & 700 / 100 \\ 
		\rowcolor{rowEven}Training Time per epoch (mins) & 5 & 2.5 & 11.5 & 23 \\ \hline
	\end{tabular}
	\caption{Hyperparameters used for the Molecular datasets. The format ``A / B`` specifies a two-step training procedure.}
	\label{tab:bg-hyperparameters}
\end{table*}

\section{Full Dimension Setting}
\label{app:molecule-generation}

\subsection{Datasets}
\label{app:bg}
\subsubsection{DW4, LJ13 and LJ55}	
We evaluate our method on three standard Boltzmann-generator benchmarks: DW4, LJ13, and LJ55. These benchmarks model the distributions of atomic systems governed by pairwise interaction potentials. DW4 consists of four particles in two-dimensional space interacting through a double-well potential, whereas LJ13 and LJ55 consist of 13 and 55 particles, respectively, in three-dimensional space interacting through the Lennard-Jones potential.

The double-well potential \(v_{\mathtt{DW}}\) and the Lennard-Jones potential \(v_{\mathtt{LJ}}\) are defined as
\begin{equation*}
	\begin{aligned}
		v_{\mathtt{DW}}(x_1, x_2) &= \frac{1}{2\tau} \left( a(d-d_0) + b(d-d_0)^2 + c(d-d_0)^4 \right), \\
		v_{\mathtt{LJ}}(x_1, x_2) &= \frac{\epsilon}{2\tau} \left[ \left(\frac{r_m}{d}\right)^{12} - 2\left(\frac{r_m}{d}\right)^6 \right],
	\end{aligned}
\end{equation*}
where \(d=\lVert x_1-x_2\rVert\) denotes the Euclidean distance between the two particles. We use the MCMC samples provided by \cite{klein2023} as training data\footnote{The datasets are available at \url{https://osf.io/srqg7/?view_only=28deeba0845546fb96d1b2f355db0da5}.}. We reserve 400k samples for validation and 500k samples for final testing. The inverse-temperature parameter \(\beta_t\) is set to \(1.0\) for all evaluations. Parameter values for both potentials are provided below:
\begin{itemize}
	\item DW4: \(a=0\), \(b=-4\), \(c=0.9\), \(d_0=4\), and \(\tau=1\).
	\item LJ13 and LJ55: \(r_m=1\), \(\epsilon=1\), and \(\tau=1\).
\end{itemize}

\subsubsection{QM9}

The QM9 dataset \cite{ramakrishnan2014} contains molecules with varying numbers of atoms, with the largest containing 29 atoms. We use the same data splits as \citet{NEURIPS2021_21b5680d}, 100K/18K/13K for training/validation/test respectively.

\subsection{Model and Hyperparameters}
\label{app:en-gnn}

We use the model architecture and hyperparameters proposed by \citet{draxler2024} for \(E\)-FFF, with two exceptions. For \textbf{QM9}, we reduce the batch size from 64 to 48 because of GPU memory constraints. In addition, we set the Adam optimizer's \(\beta_2\) parameter to 0.99 rather than 0.999, as this improves training stability in our experiments. The complete hyperparameter configurations for all datasets are provided in Table~\ref{tab:bg-hyperparameters}.

We train the \textbf{LJ55} model on a single NVIDIA RTX PRO 6000 GPU and all other models on a single NVIDIA RTX A5000 GPU. The training time per epoch is reported in Table~\ref{tab:bg-hyperparameters}.

\section{Bottleneck Setting}
\subsection{Second-order derivative considerations}
\label{app:second_order}
The models proposed in \citet{sorrenson2024} for Tabular and Pythae benchmark, exclusively use $\operatorname{ReLU}$ activations. However, because the surrogate loss introduces second-order derivatives into the computation graph, backpropagating through the resulting VJP/JVP operations requires evaluating $\operatorname{ReLU}''(x)$. Since $\operatorname{ReLU}''(x)=0$ almost everywhere and autodiff implementations typically assign a value of zero at the origin. The optimizer has access only to a simplified, locally linear representation of the network. Consequently, curvature information from the activation functions is discarded, and the gradient provides no signal about how changes in the weights affect the network's local sensitivity.

To address this limitation, we replace $\operatorname{ReLU}$ with $\operatorname{SiLU}$, which is smooth and has a generally nonzero second derivative, thereby preserving the second-order gradient pathway. This substitution further improves performance, albeit at the cost of increased VRAM usage and training time because of the greater computational expense of evaluating $\operatorname{SiLU}$ and its derivatives. We leave a systematic study of the effects of different activation functions on Jacobian-based training to future work.

For a fair comparison with existing $\operatorname{ReLU}$-based baselines, we evaluate our method using $\operatorname{ReLU}$ and report results obtained with $\operatorname{SiLU}$ separately to demonstrate the full potential of our approach.

\subsection{Tabular Data}
\subsubsection{Experimental Details}
\label{app:tabular_exp}

\begin{table}[t]
	\small
	\centering
	\setlength{\tabcolsep}{1.1mm}
	\rowcolors[]{2}{rowOdd}{rowEven}
	\begin{tabular}{lllll}
		\hline
		\rowcolor{highlight}Hyperparameter & \textbf{Power} & \textbf{Gas} & \textbf{HEPMASS} & \textbf{MiniBooNE} \\   \hline
		\begin{tabular}[t]{@{}l@{}} Latent\\dimension \end{tabular} & 3 & 2 & 10 & 21\\
		\begin{tabular}[t]{@{}l@{}} Training\\epochs \end{tabular}  & 15 & 30 & 85 & 875 \\
		\begin{tabular}[t]{@{}l@{}} Reconstruction\\ weight $\beta$ \end{tabular}   & 10 & 10 & 10 & 5 \\
		\begin{tabular}[t]{@{}l@{}} Learning\\ rate \end{tabular} {} & 0.001 & 0.001 & 0.0003 & 0.0001 \\ 
		\hline
	\end{tabular}
	\caption{Dataset-dependent hyperparameters for tabular data experiments.}
	\label{tab:tabular-data-hyperparams}
\end{table}
For the tabular-data experiments, we follow the setup of \citet{caterini2021}, using the same datasets, data splits, and latent-space dimensions. We adopt the model architecture proposed by \citet{sorrenson2024}: the encoder comprises a two-layer feed-forward network followed by a two-block ResNet, while the decoder mirrors this structure in reverse. All networks have a hidden width of 256 neurons.

We use the same training configuration, including a batch size of 512, isotropic Gaussian noise with a standard deviation of 0.01, \(K=1\) Hutchinson sample, gradient clipping with a maximum norm of 1.0, and the Adam optimizer with a one-cycle learning-rate scheduler. We do not use weight decay and set Adam's \(\beta_2\) parameter to 0.99 rather than 0.999, as we find that this improves training stability. The reconstruction weights \(\beta\) and learning rates are dataset-dependent and are listed in Table \ref{tab:tabular-data-hyperparams}. For each dataset, we perform five independent runs. Each run takes approximately 15 minutes on a single \textit{NVIDIA RTX A5000} GPU.

\subsubsection{Extended Results}
\label{sec_app:tabular_results}

We also evaluate all models using the \(\operatorname{SiLU}\) activation function. For the Power, HEPMASS, and MiniBooNE datasets, replacing \(\operatorname{ReLU}\) with \(\operatorname{SiLU}\) yields further performance improvements.

For the Gas dataset, we attempted to reproduce the results of \citet{caterini2021} using the authors' original code but obtained a best FID of 0.183, which is notably worse than the reported value. Similar reproduction discrepancies have been documented by \citet{flouris2023} and \citet{sorrenson2024}. For completeness, Table \ref{tab:tabular_results_2} also includes the Gas reproduction results reported by \citet{flouris2023}.

\begin{table}[t]
	\setlength{\tabcolsep}{1mm}
	\small
	\centering{
		\rowcolors[]{2}{rowOdd}{rowEven}
		\begin{tabular}{l|c|c|c|c}
			\hline
			\rowcolor{highlight}Method           &        \textbf{Power}        &         \textbf{Gas}         &       \textbf{HEPMASS}       &      \textbf{MiniBooNE}      \\ \hline
			M-Flow                               &        0.258$\pm$.045        &        0.219$\pm$.016        &        0.741$\pm$.052        &        1.650$\pm$.105        \\
			RF                                   &        0.083$\pm$.015        &   \textbf{0.110$\pm$.021}    &        0.779$\pm$.191        &        1.001$\pm$.051        \\
			\textbf{RF$^\ast$}                              &      0.074$\pm$.012       &      0.283$\pm$.031       &      0.628$\pm$.046       &      1.622$\pm$.121       \\
			CMF                                  &        0.053$\pm$.005        &        0.373$\pm$.065        &        0.574$\pm$.044        &        1.508$\pm$.082        \\
			FIF                                  & \underline{{0.041$\pm$.007}} &        0.281$\pm$.031        & \underline{{0.541$\pm$.034}} & \underline{{0.598$\pm$.024}} \\
			NAE                                  &   \textbf{0.013$\pm$.005}    & \underline{{0.216$\pm$.013}} &   \textbf{0.485$\pm$.025}    &   \textbf{0.583$\pm$.027}    \\
			\hiderowcolors				\hline
			NAE-\(\operatorname{S}\) &       \textit{0.011$\pm$.003}       &       {0.252$\pm$.006}       &      \textit{0.435$\pm$0.016}       &       \textit{0.534$\pm$.014}       \\ \hline
		\end{tabular}
	}
	\caption{%
		Results on tabular datasets. Average FID scores (lower is better) and standard deviations across 5 runs are reported. \textbf{Bold} indicates the best performance, while \underline{{underlined}} indicates the second-best. M-Flow and \textbf{RF$^\ast$} are the results reproduced by CMF. Results for NAE-\(\operatorname{S}\) are included solely to demonstrate the potential of our method when paired with an appropriate activation function; they are not considered for direct comparison. However, when they outperform all other models, they are marked in \textit{italics}.
	}
	\label{tab:tabular_results_2}
\end{table}

\subsection{Pythae Benchmark}

\subsubsection{Experimental Details}
\label{app:pythae_exp}

For the vision experiments, we follow the benchmark setup and training protocols of \citet{chadebec2022}. We adopt the same ConvNet and ResNet architectures used in the benchmark, as detailed in Tables \ref{tab:pythae-conv-net} and \ref{tab:pythae-res-net}. For MNIST and CIFAR-10, we train each model for 100 epochs using the Adam optimizer with an initial learning rate of \(10^{-4}\), reserving the final 10,000 images of the training set for validation. For CelebA, we train for 50 epochs with an initial learning rate of \(10^{-3}\). All experiments use a batch size of 100, and the learning rate is halved when the validation loss does not improve for 10 consecutive epochs. Following the original benchmark, we perform a hyperparameter search over 10 configurations, select the model with the best validation FID, and report its test-set FID and Inception Score. Following \citet{sorrenson2024}, we search over five reconstruction weights, \(\beta \in \{2,5,10,20,25\}\), and two numbers of Hutchinson probes, \(K \in \{1,2\}\).

Following \citet{sorrenson2024}, we exclude VAEGAN from the comparison because of its substantially higher training cost.

\begin{table*}[t]
	\centering
	\small
	\rowcolors[]{2}{rowOdd}{rowEven}
	\begin{tabular}{cccc}
		\hline
		\rowcolor{highlight}Encoder& MNIST & CIFAR10 & CELEBA \\
		\hline
		\textit{Input} & (1, 28, 28)& (3, 32, 32) & (3, 64, 64) \\
		{Layer 1} & Conv(128, 4, 2), BN, ReLU & Conv(128, 4, 2), BN, ReLU & Conv(128, 4, 2), BN, ReLU\\
		{Layer 2} & Conv(256, 4, 2), BN, ReLU & Conv(256, 4, 2), BN, ReLU & Conv(256, 4, 2), BN, ReLU\\
		{Layer 3} & Conv(512, 4, 2), BN, ReLU & Conv(512, 4, 2), BN, ReLU & Conv(512, 4, 2), BN, ReLU \\
		{Layer 4} & Conv(1024, 4, 2), BN, ReLU & Conv(1024, 4, 2), BN, ReLU & Conv(1024, 4, 2), BN, ReLU \\
		{Layer 5}& Linear(1024, 16) & Linear(4096, 256) & Linear(16384, 64)\\
		\midrule \hline
		\rowcolor{highlight}Decoder & MNIST & CIFAR10 & CELEBA  \\ \hline
		Layer 1 & Linear(16, 16384) & Linear(256, 65536) & Linear(64, 65536) \\
		\multirow{1}{*}{Layer 2} & ConvT(512, 3, 2), BN, ReLU & ConvT(512, 4, 2), BN, ReLU & ConvT(512, 5, 2), BN, ReLU \\
		\multirow{1}{*}{Layer 3} & ConvT(256, 3, 2), BN, ReLU & ConvT(256, 4, 2), BN, ReLU & ConvT(256, 5, 2), BN, ReLU \\
		\multirow{1}{*}{Layer 4} & Conv(1, 3, 2), Sigmoid & Conv(3, 4, 1), Sigmoid & ConvT(128, 5, 2), BN, ReLU \\
		{Layer 5} & - & - & ConvT(3, 5, 1), Sigmoid \\
		\midrule
		\#Parameters & 17.2M & 39.4M & 33.5M \\
		\bottomrule
	\end{tabular}
	
	\caption{{ConvNet}, neural network architecture used for the convolutional networks, adapted from \citet{sorrenson2024}.}
	
	\label{tab:pythae-conv-net}
\end{table*}

\begin{table*}[t]
	\centering
	\small
	\rowcolors[]{2}{rowOdd}{rowEven}
	\begin{tabular}{cccc}
		\hline
		\rowcolor{highlight}Encoder & MNIST & CIFAR10 & CELEBA \\
		\hline
		\textit{Input} & (1, 28, 28)& (3, 32, 32) & (3, 64, 64) \\
		\multirow{1}{*}{Layer 1} & Conv(64, 4, 2) & Conv(64, 4, 2) & Conv(64, 4, 2) \\
		\multirow{1}{*}{Layer 2} & Conv(128, 4, 2) & Conv(128, 4, 2) & Conv(128, 4, 2)\\
		\multirow{1}{*}{Layer 3} & Conv(128, 3, 2) & Conv(128, 3, 1) & Conv(128, 3, 2)\\
		\multirow{1}{*}{Layer 4} & ResBlock* & ResBlock* & Conv(128, 3, 2)\\
		\multirow{1}{*}{Layer 5} & ResBlock* & ResBlock* & ResBlock* \\
		\multirow{1}{*}{Layer 6} & Linear(2048, 16) & Linear(8192, 256) & ResBlock* \\
		\multirow{1}{*}{Layer 7}& - & - & Linear(2048, 64)\\
		\midrule \hline
		\rowcolor{highlight}Decoder & MNIST & CIFAR10 & CELEBA \\ \hline
		Layer 1 & Linear(16, 2048) & Linear(256, 8192) & Linear(64, 2048) \\
		\multirow{1}{*}{Layer 2} & ConvT(128, 3, 2) & ResBlock* & ConvT(128, 3, 2)\\ 
		\multirow{1}{*}{Layer 3} & ResBlock* & ResBlock* & ResBlock*\\
		\multirow{1}{*}{Layer 4} & ResBlock*, ReLU & ConvT(64, 4, 2) & ResBlock* \\
		\multirow{1}{*}{Layer 5} & ConvT(64, 3, 2), ReLU &  ConvT(3, 4, 2), Sigmoid & ConvT(128, 5, 2), Sigmoid \\
		\multirow{1}{*}{Layer 6} & ConvT(1, 3, 2), Sigmoid & - & ConvT(64, 5, 2), Sigmoid \\
		\multirow{1}{*}{Layer 6} & - & - & ConvT(3, 4, 2), Sigmoid \\ \hline
		\midrule
		\#Parameters & 0.73M & 4.8M & 1.6M \\
		\bottomrule
		
	\end{tabular}
	\caption{{ResNet}, neural network architecture used for the residual networks, adapted from \citet{sorrenson2024}. *The ResBlocks are composed of one Conv(32, 3, 1) followed by Conv(128, 1, 1) with ReLU.}
	
	\label{tab:pythae-res-net}
\end{table*}

\subsubsection{Extended Results}
\label{app:pythae_results}

Complete results for the Pythae benchmark are reported in Table~\ref{tab:benchmark_results}. The large ConvNet-based models exhibit signs of overfitting, particularly on MNIST and CIFAR-10. We hypothesize that this behavior is partly attributable to the absence of data augmentation and dropout in the Pythae training pipeline. For consistency with the original benchmark, however, we leave the training protocol unchanged.

\begin{table*}[!t]
	\centering
	\small
	\rowcolors[]{5}{rowOdd}{rowEven}
	\begin{tabular}{c|c|rr|rr|rr|rr|rr|rr}
		\toprule
		&                           &                                   \multicolumn{6}{c|}{ConvNet}                                    &                                 \multicolumn{6}{c}{ResNet}                                  \\
		\multirow{1}{*}{Model}                                   &                           &    \multicolumn{2}{c|}{MNIST}     & \multicolumn{2}{c|}{CIFAR10} &   \multicolumn{2}{c|}{CELEBA}    &    \multicolumn{2}{c|}{MNIST}     & \multicolumn{2}{c|}{CIFAR10} & \multicolumn{2}{c}{CELEBA} \\
		\multirow{1}{*}{(\# Params)}                                  &                           &    \multicolumn{2}{c|}{(17.2M)}     & \multicolumn{2}{c|}{(39.4M)} &   \multicolumn{2}{c|}{(33.5M)}    &    \multicolumn{2}{c|}{(0.73M)}     & \multicolumn{2}{c|}{(4.8M)} & \multicolumn{2}{c}{(1.6M)} \\
		& \multirow{-4}{*}{Sampler} & FID  & IS  &        FID        &   IS    &       FID        & IS  & FID & IS &        FID        &   IS    &       FID        &   IS    \\ \midrule
		&       $\mathcal{N}$       &       28.9       &      2.3      &       139.7       &   2.7   &       57.6       &      2.2      &       20.6       &      2.1      &       130.9       &   2.7   &  \textbf{60.4}   &   1.7   \\
		
		\multirow{-2}{*}{ NAE }            &            GMM            &       10.2       &      2.2      &       112.1       &   3.4   &  \textbf{36.9}   &      2.1      &       12.2       &      2.1      &       122.9       &   3.3   &  \textbf{53.4}   &   1.8   \\

		&       $\mathcal{N}$       &       28.3       &      2.3      &       125.2       &   3.1   &       54.2       &      2.1      &  {13.5}   &      2.1      &  \textit{104.0}  &   3.7   & \textit{56.1}  &   1.7   \\
		
		\multirow{-2}{*}{ NAE-$\operatorname{SiLU}$ } &            GMM            &       9.5        &      2.2      &       125.0       &   3.7   &       39.3       &      2.1      &       12.6       &      2.1      &  \textit{103.6}   &   4.0   & \textit{{53.3}}  &   1.8   \\ \midrule

		&       $\mathcal{N}$       &       23.8       &      2.2      &       121.0       &   3.0   &       56.9       &      2.1      &       19.5       &      2.1      &       132.6       &   2.9   &      {62.3}      &   1.7   \\
		\multirow{-2}{*}{ FIF }            &            GMM            &       11.0       &      2.2      &       90.6        &   4.0   &      {47.3}      &      1.9      &       11.7       &      2.1      &       119.2       &   3.4   &      {55.0}      &   1.8   \\ \midrule
		&       $\mathcal{N}$       &       28.5       &      2.1      &       241.0       &   2.2   & \underline{54.8} &      1.9      &       31.3       &      2.0      &       181.7       &   2.5   &       66.6       &   1.6   \\
		\multirow{-2}{*}{VAE}             &            GMM            &       26.9       &      2.1      &       235.9       &   2.3   &       52.4       &      1.9      &       32.3       &      2.1      &       179.7       &   2.5   &       63.0       &   1.7   \\ \midrule
		\multirow{1}{*}{VAMP}             &           VAMP            &       64.2       &      2.0      &       329.0       &   1.5   &       56.0       &      1.9      &       34.5       &      2.1      &       181.9       &   2.5   &       67.2       &   1.6   \\ \midrule
		&       $\mathcal{N}$       &       29.0       &      2.1      &       245.3       &   2.1   &       55.7       &      1.9      &       32.4       &      2.0      &       191.2       &   2.4   &       67.6       &   1.6   \\
		\multirow{-2}{*}{IWAE}             &            GMM            &       28.4       &      2.1      &       241.2       &   2.1   &       52.7       &      1.9      &       34.4       &      2.1      &       188.8       &   2.4   &       64.1       &   1.7   \\
		&       $\mathcal{N}$       &       29.3       &      2.1      &       240.3       &   2.1   &       56.5       &      1.9      &       32.5       &      2.0      &       185.5       &   2.4   &       67.1       &   1.6   \\
		\multirow{-2}{*}{VAE-lin NF}          &            GMM            &       28.4       &      2.1      &       237.0       &   2.2   &       53.3       &      1.9      &       33.1       &      2.1      &       183.1       &   2.5   &       62.8       &   1.7   \\
		&       $\mathcal{N}$       &       27.5       &      2.1      &       236.0       &   2.2   &       55.4       &      1.9      &       30.6       &      2.0      &       183.6       &   2.5   &       66.2       &   1.6   \\
		\multirow{-2}{*}{VAE-IAF}           &            GMM            &       27.0       &      2.1      &       235.4       &   2.2   &       53.6       &      1.9      &       32.2       &      2.1      &       180.8       &   2.5   &       62.7       &   1.7   \\ \midrule
		&       $\mathcal{N}$       &       21.4       &      2.1      &  \textbf{115.4}   &   3.6   &       56.1       &      1.9      &  \textbf{19.1}   &      2.0      &  \textbf{124.9}   &   3.4   &       65.9       &   1.6   \\
		\multirow{-2}{*}{$\beta$-VAE}         &            GMM            &       9.2        &      2.2      &       92.2        &   3.9   &       51.7       &      1.9      &       11.4       &      2.1      & \underline{112.6} &   3.6   &       59.3       &   1.7   \\ \midrule
		&    $\mathcal{N}(0,1)$     &       96.5       &      2.3      &       219.4       &   3.6   &      130.8       &      1.6      &      109.4       &      2.7      &       209.8       &   3.2   &      110.6       &   1.5   \\
		&            GMM            &      192.7       &      2.3      &       300.8       &   1.8   &       94.0       &      1.5      &       98.7       &      2.1      &       202.3       &   2.6   &       83.2       &   1.6   \\
		&       $2$-s sampler       &      236.1       &      1.5      &       371.2       &   1.0   &      167.9       &      1.0      &      250.8       &      1.1      &       349.0       &   1.2   &      161.0       &   1.2   \\
		\multirow{-4}{*}{Dis $\beta$-VAE}       &        MAF sampler        &      191.8       &      2.2      &       300.6       &   1.8   &       94.3       &      1.5      &       98.3       &      2.1      &       200.6       &   2.6   &       82.6       &   1.7   \\ \midrule
		&       $\mathcal{N}$       &       21.3       &      2.1      & \underline{116.6} &   2.8   &       55.7       &      1.8      &       20.7       &      2.0      & \underline{125.8} &   3.4   &       65.9       &   1.6   \\
		\multirow{-2}{*}{$\beta$-TC VAE}        &            GMM            &       11.6       &      2.2      & \underline{89.3}  &   4.1   &       51.8       &      1.9      &       13.3       &      2.1      &  \textbf{106.5}   &   3.7   &       59.3       &   1.7   \\
		&       $\mathcal{N}$       &       27.0       &      2.1      &       236.5       &   2.2   &  \textbf{53.8}   &      1.9      &       31.0       &      2.0      &       185.4       &   2.5   &       66.4       &   1.7   \\
		\multirow{-2}{*}{FactorVAE}          &            GMM            &       26.9       &      2.1      &       234.0       &   2.2   &       52.4       &      2.0      &       32.7       &      2.1      &       184.4       &   2.5   &       63.3       &   1.7   \\ \midrule
		&       $\mathcal{N}$       &       27.5       &      2.1      &       235.2       &   2.1   &       55.5       &      1.9      &       31.1       &      2.0      &       182.8       &   2.5   &       66.5       &   1.6   \\
		\multirow{-2}{*}{InfoVAE - RBF}        &            GMM            &       26.7       &      2.1      &       230.4       &   2.2   &       52.7       &      1.9      &       32.3       &      2.1      &       179.5       &   2.5   &       62.8       &   1.7   \\
		&       $\mathcal{N}$       &       28.3       &      2.1      &       233.8       &   2.2   &       56.7       &      1.9      &       31.0       &      2.0      &       182.4       &   2.5   &       66.4       &   1.6   \\
		\multirow{-2}{*}{InfoVAE - IMQ}        &            GMM            &       27.7       &      2.1      &       231.9       &   2.2   &       53.7       &      1.9      &       32.8       &      2.1      &       180.7       &   2.6   &       62.3       &   1.7   \\
		&       $\mathcal{N}$       &  \textbf{16.8}   &      2.2      &       139.9       &   2.6   &       59.9       &      1.8      &  \textbf{19.1}   &      2.1      &       164.9       &   2.4   & \underline{64.8} &   1.7   \\
		\multirow{-2}{*}{AAE}             &            GMM            &       9.3        &      2.2      &       92.1        &   3.8   &       53.9       &      2.0      &       11.1       &      2.1      &       118.5       &   3.5   &       58.7       &   1.8   \\ \midrule
		&       $\mathcal{N}$       &       26.7       &      2.2      &       279.9       &   1.7   &      124.3       &      1.3      &       28.0       &      2.1      &       254.2       &   1.7   &      119.0       &   1.3   \\
		\multirow{-2}{*}{MSSSIM-VAE}          &            GMM            &       27.2       &      2.2      &       279.7       &   1.7   &      124.3       &      1.3      &       28.8       &      2.1      &       253.1       &   1.7   &      119.2       &   1.3   \\
		&       $\mathcal{N}$       &       26.7       &      2.1      &       201.3       &   2.1   &      327.7       &      1.0      &      221.8       &      1.3      &       210.1       &   2.1   &      275.0       &   2.9   \\
		\multirow{-2}{*}{AE}              &            GMM            &       9.3        &      2.2      &       97.3        &   3.6   &       55.4       &      2.0      & \underline{11.0} &      2.1      &       120.7       &   3.4   & \underline{57.4} &   1.8   \\
		&       $\mathcal{N}$       &       21.2       &      2.2      &       175.1       &   2.0   &      332.6       &      1.0      &       21.2       &      2.1      &       170.2       &   2.3   &       69.4       &   1.6   \\
		\multirow{-2}{*}{WAE - RBF}          &            GMM            &       9.2        &      2.2      &       97.1        &   3.6   &       55.0       &      2.0      &       11.2       &      2.1      &       120.3       &   3.4   &       58.3       &   1.7   \\
		&       $\mathcal{N}$       & \underline{18.9} &      2.2      &       164.4       &   2.2   &       64.6       &      1.7      &       20.3       &      2.1      &       150.7       &   2.5   &       67.1       &   1.6   \\
		\multirow{-2}{*}{WAE - IMQ}          &            GMM            &   \textbf{8.6}   &      2.2      &       96.5        &   3.6   &       51.7       &      2.0      &       11.2       &      2.1      &       119.0       &   3.5   &       57.7       &   1.8   \\
		&       $\mathcal{N}$       &       28.2       &      2.0      &       152.2       &   2.0   &      306.9       &      1.0      &      170.7       &      1.6      &       195.7       &   1.9   &      140.3       &   2.2   \\
		\multirow{-2}{*}{VQVAE}            &            GMM            & \underline{9.1}  &      2.2      &       95.2        &   3.7   & \underline{51.6} &      2.0      &  \textbf{10.7}   &      2.1      &       120.1       &   3.4   &       57.9       &   1.8   \\
		&       $\mathcal{N}$       &       25.0       &      2.0      &       156.1       &   2.6   &       86.1       &      2.8      &       63.3       &      2.2      &       170.9       &   2.2   &      168.7       &   3.1   \\
		\multirow{-2}{*}{RAE - L2}           &            GMM            & \underline{9.1}  &      2.2      &   \textbf{85.3}   &   3.9   &       55.2       &      1.9      &       11.5       &      2.1      &       122.5       &   3.4   &       58.3       &   1.8   \\
		&       $\mathcal{N}$       &       27.1       &      2.1      &       196.8       &   2.1   &       86.1       &      2.4      &       61.5       &      2.2      &       229.1       &   2.0   &      201.9       &   3.1   \\
		\multirow{-2}{*}{RAE - GP}           &            GMM            &       9.7        &      2.2      &       96.3        &   3.7   &       52.5       &      1.9      &       11.4       &      2.1      &       123.3       &   3.4   &       59.0       &   1.8   \\ \midrule
		VAEGAN                     &       $\mathcal{N}$       &   \textit{8.7}   &      2.2      &       199.5       &   2.2   &  \textit{39.7}   &      1.9      &      {12.8}      &      2.2      &       198.7       &   2.2   &      122.8       &   2.0   \\
		\textit{(not compared)}            &            GMM            &   \textit{6.3}   &      2.2      &       197.5       &   2.1   &  \textit{35.6}   &      1.8      &   \textit{6.5}   &      2.2      &       188.2       &   2.6   &       84.3       &   1.7   \\ \bottomrule
	\end{tabular}
	\caption{Table taken from \cite{sorrenson2024} with our results added at the top. We report Inception Score (IS, higher is better) and Fréchet Inception Distance (FID, lower is better) computed with 10k samples on the test set. The best model per dataset and sampler is highlighted in \textbf{bold}, the second best is \underline{underlined}. Results for NAE-\(\operatorname{SiLU}\) are included solely to demonstrate the potential of our method when paired with an appropriate activation function; they are not considered for direct comparison. However, when they outperform all other models, they are marked in \textit{italics}.}
	\label{tab:benchmark_results}
\end{table*}

\paragraph{Alignment between Prior and Latent distribution.}
We also evaluate the models using the \(\operatorname{SiLU}\) activation function and find that it further improves the performance of the ResNet-based architectures. Notably, for the ResNet-based NAE models with \(\operatorname{SiLU}\), sampling from the standard normal prior, \(\mathcal{N}\), and from a 10-component GMM fitted to the encoded training data produces similar FID scores.

In generative models, fitting a GMM can improve sample quality when the learned latent distribution contains disconnected, multimodal, or otherwise non-Gaussian structure that cannot be adequately represented by a unimodal standard normal distribution. A substantial reduction in FID when switching from \(\mathcal{N}\) to the fitted GMM may therefore indicate that the standard Gaussian fails to capture relevant structure in the learned latent distribution.

By contrast, the ResNet-based NAE models with \(\operatorname{SiLU}\) obtain only a small reduction in FID when samples from \(\mathcal{N}\) are replaced by samples from the fitted GMM. This limited benefit suggests that the learned latent representation is sufficiently regular and compatible with the assumed Gaussian prior, such that the additional flexibility of the GMM provides little practical advantage for generation. Together with the strong absolute performance obtained using \(\mathcal{N}\), the small FID gap therefore suggests that NAE learns a coherent latent representation that can be sampled effectively using a standard normal distribution.

Nevertheless, this result provides only indirect evidence of prior--latent alignment and does not establish that the aggregated latent distribution is Gaussian. Similar FID scores may be observed even when deviations from Gaussianity are present but have a limited effect on the generated samples or are not adequately captured by FID. A direct characterization of the aggregated latent distribution and its alignment with the assumed prior is therefore left for future work.

\section{Injective Models}

\begin{table*}[!t]
	\centering
	\small
	\rowcolors[]{2}{rowOdd}{rowEven}
	\begin{tabular}{ccc}
		\hline
		\rowcolor{highlight}Encoder & NAE & FIF \\ \hline
		\textit{Input}           & (3, 64, 64)                                          & (3, 64, 64)                             \\
		Layer 1            &  Conv(64, 4, 2), BN, SiLU                          & Conv(128, 4, 2), BN, ReLU               \\
		Layer 2            & Conv(128, 4, 2), BN, SiLU                       & Conv(256, 4, 2), BN, ReLU               \\
		Layer 3            & Conv(256, 4, 2), BN, SiLU                   & Conv(512, 4, 2), BN, ReLU               \\
		Layer 4            & Conv(512, 4, 2), BN, SiLU                       & Conv(1024, 4, 2), BN, ReLU              \\
		Layer 5            & Linear(8192, 64)                  & Linear(16384, 64)             \\
		Layer 6-9 & 4$\times$ResBlock(128)* & 4$\times$ResBlock(256)* \\ 
		\midrule \hline
		
		\rowcolor{highlight}	Decoder          & NAE & FIF                        \\ \hline
		{Layer 1-4} & {4$\times$ResBlock(128)*} & {4$\times$ResBlock(256)*} \\
		Layer 5            & Linear(64, 32768)                        & Linear(64, 65536)              \\
		Layer 6            & ConvT(256, 5, 2), BN, SiLU                        & ConvT(512, 5, 2), BN, ReLU              \\
		Layer 7            & ConvT(128, 5, 2), BN, SiLU                         & ConvT(256, 5, 2), BN, ReLU              \\
		Layer 8            & ConvT(64, 5, 2), BN, SiLU                 & ConvT(128, 5, 2), BN, ReLU              \\
		Layer 9            & ConvT(3, 5, 1), Sigmoid                         & ConvT(3, 5, 1), Sigmoid                \\
		\midrule
		\# Params & 9.9M & 34.3M \\
		\bottomrule
	\end{tabular}
	\caption{{ConvNet} NAE and FIF \cite{sorrenson2024} models used for Celeba dataset for comparison to Trumpet and Denoising Normalizing Flow. *The ResBlocks(inner\_dim) for both models are composed of Linear(latent\_dim, inner\_dim), \(\operatorname{SiLU}\), Linear(inner\_dim, inner\_dim), \(\operatorname{SiLU}\), Linear(inner\_dim, latent\_dim) with a skip connection.}
	
	\label{tab:injective_architecture}
\end{table*}

\subsubsection{Experimental Details}
\label{app:injective_exp}
We follow the experimental setup of FIF \citep{sorrenson2024} for injective flows on CelebA. Rather than rerunning Trumpets \citep{kothari2021} and Denoising Normalizing Flows (DNF) \citep{horvat2021}, we use the results reported by \citet{sorrenson2024}. The experiments in \citet{sorrenson2024} use a five-hour training budget. To account for differences in hardware, we reproduce FIF on our hardware and obtain comparable performance after approximately 100 epochs, corresponding to 6.5 hours on a single \textit{NVIDIA RTX A5000} GPU. We therefore adopt 6.5 hours as the wall-clock training budget for our method.

For the Normalizing Autoencoder (NAE), we initially use the same architecture as FIF but observe overfitting. To mitigate this, we halve the hidden dimensions of both the encoder and decoder and replace the activation function with \(\operatorname{SiLU}\). We train the model for 100 epochs using the Adam optimizer with a learning rate of \(10^{-3}\), \(\beta_1=0.9\), \(\beta_2=0.99\), a batch size of 256, and a gradient-clipping threshold of 1.0. Architecture specifications for both FIF and NAE are provided in Table~\ref{tab:injective_architecture}.

\section{Tabular Ablation}

\subsubsection{Experimental Details}
\label{app:ablation_exp}

We perform a controlled ablation study on the Power dataset to characterize the sensitivity of the Conditional loss to the latent dimension \(d\) and reconstruction weight \(\beta\). \(d\) determines the dimensionality of the learned representation, whereas \(\beta\) controls the relative contribution of the reconstruction term. Varying them jointly allows us to assess how the balance between likelihood optimization and approximate inversion changes with latent dimensionality and to compare this behaviour with that of the Encoder Surrogate loss.

\begin{figure}[!b]
	\centering
	\includegraphics[width=0.34\textwidth]{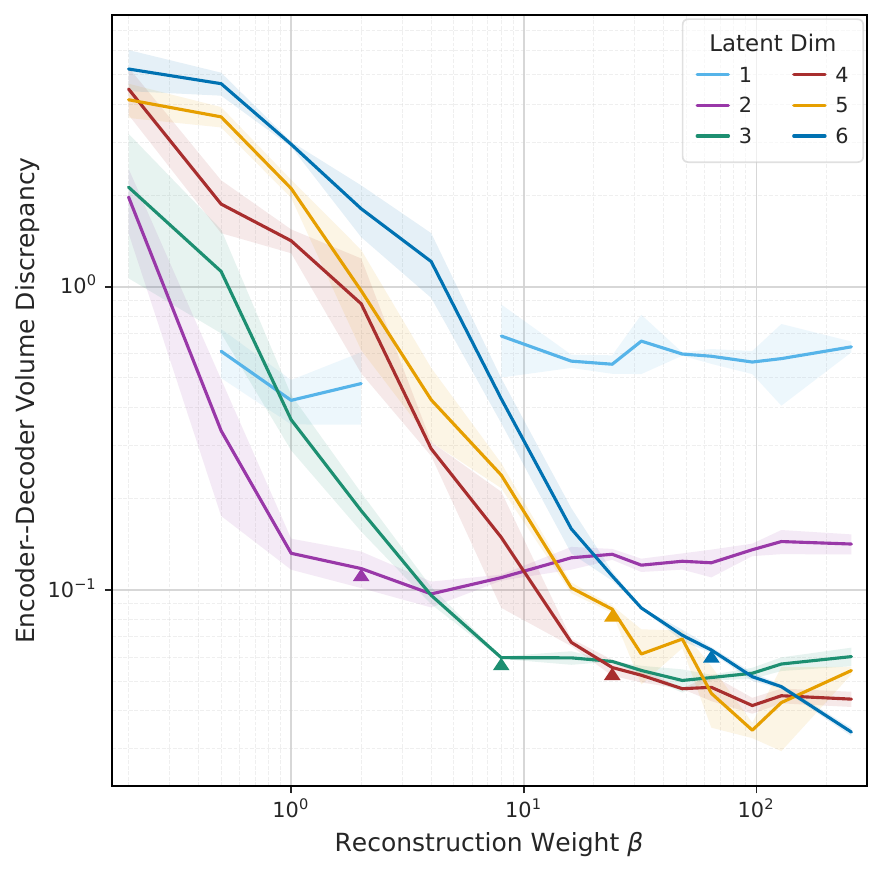}
	\caption{Encoder-decoder volume discrepancy for the Conditional loss.
		For each latent dimension \(d\) and reconstruction weight \(\beta\), we compute the volume change of the encoder and decoder on the held-out test set using exact Jacobians. We report their absolute discrepancy normalized by the latent dimension, \(\Delta_{\mathrm{vol}}/d\), for the final model. Curves show the mean over three independent runs, with colours indicating the latent dimension. Lower values indicate greater local geometric consistency between the encoder and decoder. Shaded regions indicate one standard deviation, and triangular markers denote the reconstruction weight that attains the lowest mean FID-like score for each latent dimension.
	}
	\label{fig:conditional_volume_discrepancy}
\end{figure}

We use the same model architecture, data preprocessing, optimizer, and training configuration as in the main tabular experiments, with \(\operatorname{ReLU}\) activations. We sweep over all latent dimensions $d \in \{1,2,3,4,5,6\}$ and reconstruction weights:
\[
\beta \in \{0.2,0.5,1,2,4,8,16,24,32,48,64,96,128,256\}.
\]
For each pair \((d,\beta)\), we train both objectives using three independent random seeds and evaluate them under the same protocol. Each experiment takes approximately 15 minutes on a single \textit{NVIDIA RTX A5000} GPU.

\subsubsection{Extended Results}
\label{app:ablation_results}

To examine the approximate-inverse regime more directly, we measure the consistency between the local volume changes induced by the encoder and decoder. Specifically, we compute both quantities from exact Jacobians on the test set and report their absolute discrepancy normalized by the latent dimension. If the encoder and decoder are approximate inverses, their local geometric descriptions should be consistent, resulting in a small volume discrepancy.

As shown in Fig.~\ref{fig:conditional_volume_discrepancy}, increasing the reconstruction weight generally reduces the volume discrepancy. For \(d \geq 2\), the onset of the approximate-inverse regime identified by the reconstruction diagnostics coincides with the normalized discrepancy decreasing to approximately \(10^{-1}\) or below. We interpret this value as an empirical diagnostic rather than a universal threshold, as a small volume discrepancy alone is insufficient to establish approximate inversion. For \(d=1\), the behaviour is erratic.

\newpage

\bibliography{Zotero_Library}

@misc{dinh2017,
  title = {Density Estimation Using {{Real NVP}}},
  author = {Dinh, Laurent and {Sohl-Dickstein}, Jascha and Bengio, Samy},
  year = 2016,
  month = may,
  number = {arXiv:1605.08803},
  eprint = {1605.08803},
  primaryclass = {cs},
  publisher = {arXiv},
  doi = {10.48550/arXiv.1605.08803},
  urldate = {2025-02-18},
  archiveprefix = {arXiv}
}

@article{grathwohl2018,
  title = {{{FFJORD}}: {{Free-form Continuous Dynamics}} for {{Scalable Reversible Generative Models}}},
  shorttitle = {{{FFJORD}}},
  author = {Grathwohl, Will and Chen, Ricky T. Q. and Bettencourt, Jesse and Sutskever, Ilya and Duvenaud, David},
  year = 2018,
  journal = {International Conference on Learning Representations},
  eprint = {1810.01367},
  primaryclass = {cs},
  publisher = {arXiv},
  doi = {10.48550/arXiv.1810.01367},
  urldate = {2025-02-18},
  archiveprefix = {arXiv}
}

@misc{kingma2018,
  title = {Glow: {{Generative Flow}} with {{Invertible}} 1x1 {{Convolutions}}},
  shorttitle = {Glow},
  author = {Kingma, Diederik P. and Dhariwal, Prafulla},
  year = 2018,
  month = jul,
  number = {arXiv:1807.03039},
  eprint = {1807.03039},
  primaryclass = {stat},
  publisher = {arXiv},
  doi = {10.48550/arXiv.1807.03039},
  urldate = {2025-06-26},
  archiveprefix = {arXiv}
}

@misc{chen2020,
  title = {Residual {{Flows}} for {{Invertible Generative Modeling}}},
  author = {Chen, Ricky T. Q. and Behrmann, Jens and Duvenaud, David and Jacobsen, J{\"o}rn-Henrik},
  year = 2020,
  month = jul,
  number = {arXiv:1906.02735},
  eprint = {1906.02735},
  primaryclass = {stat},
  publisher = {arXiv},
  doi = {10.48550/arXiv.1906.02735},
  urldate = {2025-09-03},
  archiveprefix = {arXiv}
}

@inproceedings{dinh2015,
  title = {{{NICE}}: Non-Linear Independent Components Estimation},
  booktitle = {3rd International Conference on Learning Representations, {{ICLR}} 2015, San Diego, {{CA}}, {{USA}}, May 7-9, 2015, Workshop Track Proceedings},
  author = {Dinh, Laurent and Krueger, David and Bengio, Yoshua},
  editor = {Bengio, Yoshua and LeCun, Yann},
  year = 2015,
  eprint = {1410.8516},
  primaryclass = {cs.LG},
  publisher = {arXiv},
  archiveprefix = {arXiv},
  bibsource = {dblp computer science bibliography, https://dblp.org}
}

@misc{ho2019,
  title = {Flow++: {{Improving Flow-Based Generative Models}} with {{Variational Dequantization}} and {{Architecture Design}}},
  shorttitle = {Flow++},
  author = {Ho, Jonathan and Chen, Xi and Srinivas, Aravind and Duan, Yan and Abbeel, Pieter},
  year = 2019,
  month = may,
  number = {arXiv:1902.00275},
  eprint = {1902.00275},
  primaryclass = {cs},
  publisher = {arXiv},
  doi = {10.48550/arXiv.1902.00275},
  urldate = {2025-09-18},
  archiveprefix = {arXiv}
}

@misc{durkan2019,
  title = {Cubic-{{Spline Flows}}},
  author = {Durkan, Conor and Bekasov, Artur and Murray, Iain and Papamakarios, George},
  year = 2019,
  month = jun,
  number = {arXiv:1906.02145},
  eprint = {1906.02145},
  primaryclass = {stat},
  publisher = {arXiv},
  doi = {10.48550/arXiv.1906.02145},
  urldate = {2025-09-28},
  archiveprefix = {arXiv}
}

@inproceedings{walton2023,
  title = {Isomorphism, {{Normalizing Flows}}, and {{Density Estimation}}: {{Preserving Relationships Between Data}}},
  shorttitle = {Isomorphism, {{Normalizing Flows}}, and {{Density Estimation}}},
  author = {Walton, Steven},
  year = 2023,
  url={https://api.semanticscholar.org/CorpusID:261404896},
  volume = {Technical Report},
  urldate = {2025-10-14}
}

@misc{kothe2023,
  title = {A {{Review}} of {{Change}} of {{Variable Formulas}} for {{Generative Modeling}}},
  author = {K{\"o}the, Ullrich},
  year = 2023,
  number = {arXiv:2308.02652},
  eprint = {2308.02652},
  primaryclass = {cs},
  publisher = {arXiv},
  doi = {10.48550/arxiv.2308.02652},
  urldate = {2025-10-14},
  archiveprefix = {arXiv}
}

@inproceedings{draxler2024,
  title = {Free-Form {{Flows}}: {{Make Any Architecture}} a {{Normalizing Flow}}},
  shorttitle = {Free-Form {{Flows}}},
  booktitle = {Proceedings of {{The}} 27th {{International Conference}} on {{Artificial Intelligence}} and {{Statistics}}},
  author = {Draxler, Felix and Sorrenson, Peter and Zimmermann, Lea and Rousselot, Armand and K{\"o}the, Ullrich},
  year = 2024,
  month = apr,
  pages = {2197--2205},
  publisher = {PMLR},
  issn = {2640-3498},
  urldate = {2026-07-14},
  langid = {english}
}

@book{krantz2008,
  title = {Geometric {{Integration Theory}}},
  author = {Krantz, Steven G and Parks, Harold R},
  year = 2008,
  month = dec,
  publisher = {Birkh\"auser Boston},
  address = {Boston},
  doi = {10.1007/978-0-8176-4679-0},
  isbn = {978-0-8176-4676-9 978-0-8176-4679-0},
  langid = {english}
}

@misc{cramer2023,
  title = {Nonlinear {{Isometric Manifold Learning}} for {{Injective Normalizing Flows}}},
  author = {Cramer, Eike and Rauh, Felix and Mitsos, Alexander and Tempone, Ra{\'u}l and Dahmen, Manuel},
  year = 2023,
  month = may,
  number = {arXiv:2203.03934},
  eprint = {2203.03934},
  primaryclass = {cs},
  publisher = {arXiv},
  doi = {10.48550/arXiv.2203.03934},
  urldate = {2026-01-11},
  archiveprefix = {arXiv}
}

@misc{ross2021,
  title = {Tractable {{Density Estimation}} on {{Learned Manifolds}} with {{Conformal Embedding Flows}}},
  author = {Ross, Brendan Leigh and Cresswell, Jesse C.},
  year = 2021,
  month = nov,
  number = {arXiv:2106.05275},
  eprint = {2106.05275},
  primaryclass = {stat},
  publisher = {arXiv},
  doi = {10.48550/arXiv.2106.05275},
  urldate = {2026-01-11},
  archiveprefix = {arXiv}
}

@misc{kim2020,
  title = {{{SoftFlow}}: {{Probabilistic Framework}} for {{Normalizing Flow}} on {{Manifolds}}},
  shorttitle = {{{SoftFlow}}},
  author = {Kim, Hyeongju and Lee, Hyeonseung and Kang, Woo Hyun and Lee, Joun Yeop and Kim, Nam Soo},
  year = 2020,
  month = nov,
  number = {arXiv:2006.04604},
  eprint = {2006.04604},
  primaryclass = {cs},
  publisher = {arXiv},
  doi = {10.48550/arXiv.2006.04604},
  urldate = {2026-01-14},
  archiveprefix = {arXiv}
}

@inproceedings{horvat2021,
  title = {Denoising {{Normalizing Flow}}},
  booktitle = {Advances in {{Neural Information Processing Systems}}},
  author = {Horvat, Christian and Pfister, Jean-Pascal},
  year = 2021,
  volume = {34},
  pages = {9099--9111},
  publisher = {Curran Associates, Inc.},
  urldate = {2026-01-14}
}

@article{sorrenson2024,
  title = {Lifting {{Architectural Constraints}} of {{Injective Flows}}},
  author = {Sorrenson, Peter and Draxler, Felix and Rousselot, Armand and Hummerich, Sander and Zimmermann, Lea and K{\"o}the, Ullrich},
  year = 2023,
  journal = {International Conference on Learning Representations},
  eprint = {2306.01843},
  primaryclass = {cs},
  publisher = {arXiv},
  doi = {10.48550/arXiv.2306.01843},
  urldate = {2026-01-14},
  archiveprefix = {arXiv}
}

@inproceedings{sorrenson2024b,
  title = {Learning {{Distributions}} on {{Manifolds}} with {{Free-Form Flows}}},
  booktitle = {Advances in {{Neural Information Processing Systems}}},
  author = {Sorrenson, Peter and Draxler, Felix and Rousselot, Armand and Hummerich, Sander and K{\"o}the, Ullrich},
  year = 2024,
  volume = {37},
  pages = {66961--66994},
  publisher = {Curran Associates, Inc.},
  address = {Vancouver, BC, Canada},
  doi = {10.52202/079017-2138},
  urldate = {2026-07-14},
  isbn = {979-8-3313-1438-5}
}

@inproceedings{hoogeboom2022a,
  title = {Equivariant {{Diffusion}} for {{Molecule Generation}} in {{3D}}},
  booktitle = {Proceedings of the 39th {{International Conference}} on {{Machine Learning}}},
  author = {Hoogeboom, Emiel and Satorras, V{\'\i}ctor Garcia and Vignac, Cl{\'e}ment and Welling, Max},
  year = 2022,
  month = jun,
  eprint = {2203.17003},
  primaryclass = {cs.LG},
  pages = {8867--8887},
  publisher = {PMLR},
  issn = {2640-3498},
  urldate = {2026-07-14},
  archiveprefix = {arXiv},
  langid = {english}
}

@misc{satorras2022a,
  title = {E(n) {{Equivariant Graph Neural Networks}}},
  author = {Satorras, Victor Garcia and Hoogeboom, Emiel and Welling, Max},
  year = 2022,
  month = feb,
  number = {arXiv:2102.09844},
  eprint = {2102.09844},
  primaryclass = {cs.LG},
  publisher = {arXiv},
  doi = {10.48550/arXiv.2102.09844},
  urldate = {2026-06-25},
  archiveprefix = {arXiv}
}

@misc{bohm2022a,
  title = {Probabilistic {{Autoencoder}}},
  author = {B{\"o}hm, Vanessa and Seljak, Uro{\v s}},
  year = 2020,
  number = {arXiv:2006.05479},
  eprint = {2006.05479},
  primaryclass = {cs.LG},
  publisher = {arXiv},
  doi = {10.48550/arXiv.2006.05479},
  urldate = {2026-07-05},
  archiveprefix = {arXiv}
}

@misc{beitler2021,
  title = {{{PIE}}: {{Pseudo-Invertible Encoder}}},
  shorttitle = {{{PIE}}},
  author = {Beitler, Jan Jetze and Sosnovik, Ivan and Smeulders, Arnold},
  year = 2021,
  month = oct,
  number = {arXiv:2111.00619},
  eprint = {2111.00619},
  primaryclass = {cs.LG},
  publisher = {arXiv},
  doi = {10.48550/arXiv.2111.00619},
  urldate = {2026-07-05},
  archiveprefix = {arXiv}
}

@misc{kohler2020,
  title = {Equivariant {{Flows}}: {{Exact Likelihood Generative Learning}} for {{Symmetric Densities}}},
  shorttitle = {Equivariant {{Flows}}},
  author = {K{\"o}hler, Jonas and Klein, Leon and No{\'e}, Frank},
  year = 2020,
  month = oct,
  number = {arXiv:2006.02425},
  eprint = {2006.02425},
  primaryclass = {stat.ML},
  publisher = {arXiv},
  doi = {10.48550/arXiv.2006.02425},
  urldate = {2026-07-12},
  archiveprefix = {arXiv}
}

@article{ramakrishnan2014,
  title = {Quantum Chemistry Structures and Properties of 134 Kilo Molecules},
  author = {Ramakrishnan, Raghunathan and Dral, Pavlo O. and Rupp, Matthias and {von Lilienfeld}, O. Anatole},
  year = 2014,
  month = aug,
  journal = {Scientific Data},
  volume = {1},
  number = {1},
  pages = {140022},
  publisher = {Nature Publishing Group},
  issn = {2052-4463},
  doi = {10.1038/sdata.2014.22},
  urldate = {2026-07-13},
  copyright = {2014 The Author(s)},
  langid = {english}
}

@inproceedings{flouris2023,
  title = {Canonical Normalizing Flows for Manifold Learning},
  booktitle = {Advances in {{Neural Information Processing Systems}}},
  author = {Flouris, Kyriakos and Konukoglu, Ender},
  year = 2023,
  volume = {36},
  pages = {27294--27314},
  publisher = {Curran Associates, Inc.},
  address = {New Orleans, Louisiana, USA},
  doi = {10.52202/075280-1189},
  urldate = {2026-07-14},
  isbn = {978-1-7138-9911-2}
}

@inproceedings{brehmer2020,
  title = {Flows for Simultaneous Manifold Learning and Density Estimation},
  booktitle = {Advances in {{Neural Information Processing Systems}}},
  author = {Brehmer, Johann and Cranmer, Kyle},
  year = 2020,
  volume = {33},
  pages = {442--453},
  publisher = {Curran Associates, Inc.},
  urldate = {2026-07-14}
}

@inproceedings{behrmann2019,
  title = {Invertible {{Residual Networks}}},
  booktitle = {Proceedings of the 36th {{International Conference}} on {{Machine Learning}}},
  author = {Behrmann, Jens and Grathwohl, Will and Chen, Ricky T. Q. and Duvenaud, David and Jacobsen, Joern-Henrik},
  year = 2019,
  month = may,
  pages = {573--582},
  publisher = {PMLR},
  issn = {2640-3498},
  urldate = {2026-07-14},
  langid = {english}
}

@inproceedings{chadebec2022,
  title = {Pythae: {{Unifying Generative Autoencoders}} in {{Python}} - {{A Benchmarking Use Case}}},
  shorttitle = {Pythae},
  booktitle = {Advances in {{Neural Information Processing Systems}} 35},
  author = {Chadebec, Cl{\'e}ment and Vincent, Louis and Allassonniere, Stephanie},
  year = 2022,
  pages = {21575--21589},
  publisher = {Neural Information Processing Systems Foundation, Inc. (NeurIPS)},
  address = {New Orleans, Louisiana, USA},
  doi = {10.52202/068431-1568},
  urldate = {2026-07-14},
  isbn = {978-1-7138-7108-8}
}

@inproceedings{caterini2021,
  title = {Rectangular {{Flows}} for {{Manifold Learning}}},
  booktitle = {Advances in {{Neural Information Processing Systems}}},
  author = {Caterini, Anthony L and {Loaiza-Ganem}, Gabriel and Pleiss, Geoff and Cunningham, John},
  year = 2021,
  volume = {34},
  pages = {30228--30241},
  publisher = {Curran Associates, Inc.},
  urldate = {2026-07-14}
}

@inproceedings{kothari2021,
  title = {Trumpets: {{Injective}} Flows for Inference and Inverse Problems},
  shorttitle = {Trumpets},
  booktitle = {Proceedings of the {{Thirty-Seventh Conference}} on {{Uncertainty}} in {{Artificial Intelligence}}},
  author = {Kothari, Konik and Khorashadizadeh, AmirEhsan and de Hoop, Maarten and Dokmani{\'c}, Ivan},
  year = 2021,
  month = dec,
  pages = {1269--1278},
  publisher = {PMLR},
  issn = {2640-3498},
  urldate = {2026-07-14},
  langid = {english}
}

@inproceedings{papamakarios2017a,
  title = {Masked {{Autoregressive Flow}} for {{Density Estimation}}},
  booktitle = {Advances in {{Neural Information Processing Systems}}},
  author = {Papamakarios, George and Pavlakou, Theo and Murray, Iain},
  year = 2017,
  volume = {30},
  publisher = {Curran Associates, Inc.},
  urldate = {2026-07-14}
}

@inproceedings{klein2023,
  title = {Equivariant Flow Matching},
  booktitle = {Advances in {{Neural Information Processing Systems}} 36},
  author = {Klein, Leon and Kr{\"a}mer, Andreas and Noe, Frank},
  year = 2023,
  pages = {59886--59910},
  publisher = {Neural Information Processing Systems Foundation, Inc. (NeurIPS)},
  address = {New Orleans, Louisiana, USA},
  doi = {10.52202/075280-2617},
  urldate = {2026-07-14},
  isbn = {978-1-7138-9911-2}
}

@article{hong2023,
  title = {Neural {{Diffeomorphic Non-uniform B-spline Flows}}},
  author = {Hong, Seongmin and Chun, Se Young},
  year = 2023,
  month = jun,
  journal = {Proceedings of the AAAI Conference on Artificial Intelligence},
  volume = {37},
  number = {10},
  pages = {12225--12233},
  issn = {2374-3468, 2159-5399},
  doi = {10.1609/aaai.v37i10.26441},
  urldate = {2026-07-14}
}

@inproceedings{NEURIPS2021_21b5680d,
  title = {E(n) Equivariant Normalizing Flows},
  booktitle = {Advances in Neural Information Processing Systems},
  author = {Garcia Satorras, Victor and Hoogeboom, Emiel and Fuchs, Fabian and Posner, Ingmar and Welling, Max},
  editor = {Ranzato, M. and Beygelzimer, A. and Dauphin, Y. and Liang, P.S. and Vaughan, J. Wortman},
  year = 2021,
  volume = {34},
  eprint = {2105.09016},
  primaryclass = {cs},
  pages = {4181--4192},
  publisher = {Curran Associates, Inc.},
  archiveprefix = {arXiv}
}

@misc{bengio2014,
  title = {Representation {{Learning}}: {{A Review}} and {{New Perspectives}}},
  shorttitle = {Representation {{Learning}}},
  author = {Bengio, Yoshua and Courville, Aaron and Vincent, Pascal},
  year = 2014,
  month = apr,
  number = {arXiv:1206.5538},
  eprint = {1206.5538},
  primaryclass = {cs.LG},
  publisher = {arXiv},
  doi = {10.48550/arXiv.1206.5538},
  urldate = {2026-07-16},
  archiveprefix = {arXiv}
}

@misc{silvestri2023,
  title = {Deterministic Training of Generative Autoencoders Using Invertible Layers},
  author = {Silvestri, Gianluigi and Roos, Daan and Ambrogioni, Luca},
  year = 2023,
  month = mar,
  number = {arXiv:2205.09546},
  eprint = {2205.09546},
  primaryclass = {stat.ML},
  publisher = {arXiv},
  doi = {10.48550/arXiv.2205.09546},
  urldate = {2026-07-20},
  archiveprefix = {arXiv}
}

\end{document}